\documentclass[graybox]{svmult}

\usepackage{mathptmx}       
\usepackage{helvet}         
\usepackage{courier}        
\usepackage{type1cm}        
\usepackage{makeidx}         
\usepackage{graphicx}        
\usepackage{multicol}        
\usepackage[bottom]{footmisc}
\usepackage{url}
\usepackage{caption}
\usepackage{subfig}

\usepackage{amssymb}
\usepackage{pdflscape}
\usepackage{multirow}
\usepackage{array}
\usepackage{url}
\usepackage{amsmath}
\usepackage{amssymb}
\usepackage{textcomp}

\makeindex             

\newcommand{\etal}{\mbox{\emph{et al.\ }}}
\newcommand{\ie}{\mbox{\emph{i.e.,\ }}}
\newcommand{\eg}{\mbox{\emph{e.g.,\ }}}

\begin{document}

\title*{Iris recognition in cases of eye pathology}

\author{Mateusz Trokielewicz, Adam Czajka, Piotr Maciejewicz}

\institute{\textbf{Mateusz Trokielewicz and Adam Czajka} \at Biometrics Laboratory, Research and Academic Computer Network (NASK)\\ Kolska 12, 01-045 Warsaw, Poland\\ Institute of Control and Computation Engineering, Warsaw University of Technology\\Nowowiejska 15/19, 00-665 Warsaw, Poland \\\email{{mateusz.trokielewicz, adam.czajka}@nask.pl}
\and \textbf{Piotr Maciejewicz} \at Department of Ophthalmology, Medical University of Warsaw\\Lindleya 4, 02-005 Warsaw, Poland\\ \email{piotr.maciejewicz@wum.edu.pl}}

\maketitle

\abstract{This chapter provides insight on how iris recognition, one of the leading biometric identification technologies in the world, can be impacted by pathologies and illnesses present in the eye, what are the possible repercussions of this influence, and what are the possible means for taking such effects into account when matching iris samples. \newline\indent
To make this study possible, a special database of iris images has been used, representing more than 20 different medical conditions of the ocular region (including cataract, glaucoma, rubeosis iridis, synechiae, iris defects, corneal pathologies and other) and containing almost 3000 samples collected from 230 distinct irises. Then, with the use of four different iris recognition methods, a series of experiments has been conducted, concluding in several important observations. \newline\indent
One of the most popular ocular disorders worldwide - the cataract - is shown to worsen genuine comparison scores when results obtained from cataract-affected eyes are compared to those coming from healthy irises. An analysis devoted to different types of impact on eye structures caused by diseases is also carried out with significant results. The enrollment process is highly sensitive to those eye conditions that make the iris obstructed or introduce geometrical distortions. Disorders affecting iris geometry, or producing obstructions are exceptionally capable of degrading the genuine comparison scores, so that the performance of the entire biometric system can be influenced. Experiments also reveal that imperfect execution of the image segmentation stage is the most prominent contributor to recognition errors.}

\section{Introduction} 
\label{sec:Introduction}
\subsection{Iris recognition}

Automatic iris recognition has emerged as an important biometric identity recognition method more than two decades ago, although the concept of recognizing people by their irises is known to have a history of 80 years. Frank Burch was the first to suggest that iris texture, and not the eye color, can be an important human identifier \cite{DaugmanAS}. This idea was later reproduced in ophthalmology textbooks to finally fall upon a breeding ground in 1993, when the first, automatic iris recognition method based on 2D Gabor wavelets was proposed by John Daugman \cite{Daugman1993}. Numerous algorithms were proposed to date, most of them inspired by the original Daugman's invention. A common pipeline of these systems begins with image acquisition in near infrared light, which is used to make this process comfortable for users, as near infrared light is almost invisible to humans. Since this kind of illumination is hardly absorbed by melanin pigment present in the iris, as opposed to visible light,  the iris texture information can be efficiently extracted even for highly pigmented, dark irises. Occlusions such as eyelids, eyelashes and specular reflections are automatically detected. The remaining, non-occluded image of iris texture is processed with various filters (\eg 2D Gabor) and filtering results are quantized to two values depending only on their signs, ending up with a binary code as a feature vector. Since iris images are normalized prior to filtering, the iris codes have identical structure for all irises. This makes it possible to use the exclusive disjunction (XOR) to calculate the number of disagreeing bits, and hence the dissimilarity score at a very high speed. While false rejections in iris recognition are mostly due to variations in the acquisition process, impostor score distributions are not influenced by such factors making the probability of false acceptance to be predictable and relatively low. Good surveys of various iris recognition methods and important research problems related to iris recognition may be found in Bowyer \etal \cite{IrisHandbook1,IrisHandbook2,BowyerSurvey2008,BowyerSurvey2010}.

The iris, placed anteriorly in the human eye and protected by the cornea, consists of a stroma, which holds a fibrovascular mesh, and of a layer of interworking muscles, whose role is to control the amount of light getting into the eyeball. The iris is relatively easy to be observed and measured. Only the structural layout of the iris' trabecular meshwork is analyzed for the purpose of extracting individual features. Neither the color, nor other iris global features (such as tissue density) are used in biometric recognition. It is believed that high degree of structural richness and uniqueness observed in iris patterns are caused by limited dependence on human genotype (a.k.a. low `genetic penetrance'). Consequently, iris recognition is often envisioned as almost `ideal biometrics', characterized by low error rates, robustness against variations in time, immunity to diseases, resistance to forgeries, and neutrality in terms of social, religious and ethical aspects. Numerous large-scale installments, such as NEXUS program \cite{CBSA}, which offers dedicated processing lanes to pre-screened travelers when entering the United States and Canada, or AADHAAR \cite{AADHAAR}, which applies biometrics for de-duplication of unique personal IDs in India, and evaluation programs such as IREX \cite{IREXgeneral}, focused on interoperability of iris recognition, present its high usability in operational scenarios.

\begin{svgraybox}
	More than two decades of operational practice and experience allowed to identify new research challenges in iris recognition. Due to successful spoofing attacks targeted at commercial iris sensors, numerous liveness detection methods were proposed at the sensor and software levels \cite{CzajkaPupilDynamicsTIFS}. Temporal stability of iris patterns seems to be lower than initially believed \cite{Fenker2011,Fenker2012,Fenker2013,Czajka2013,TrokielewiczAgingIWBF2015} and hard to be estimated as we may observe even contradictory conclusions drawn from the same datasets \cite{BowyerOrtiz2015,Grother2015}. New research in post-mortem iris recognition \cite{TrokielewiczPostMortemICB2016,TrokielewiczPostMortemBTAS2016} reveals that it is possible to use this biometric characteristic in forensic applications. This chapter is dedicated to yet another important aspect of iris recognition reliability and presents how its reliability is influenced by different eye diseases. 
\end{svgraybox}

\subsection{Possible influence of ocular disorders}
In the ISO/IEC 29794-6 standard that provides requirements for iris image quality, two possible scenarios of eye disease impact on iris recognition are distinguished \cite{ISO2}:

\begin{description}[Scenario 1]
	\item [\textbf{Scenario 1}]{in which injury or illness occurs when the affected person is already using a biometric system as a registered user (\eg a medical procedure affecting the eye is performed or an illness influencing the iris takes place); in this case, we may observe a degradation in recognition system's performance when pre-disease images are compared to those obtained afterwards;}
	\item [\textbf{Scenario 2}]{in which injury or illness is present before the enrollment and therefore an overall performance of a biometric system may be worse than when presented with healthy eyes; in severe cases the eye may not be suitable for iris recognition at all (\eg person suffers from a congenital disease called aniridia, in which only the iris tissue is absent, often leaving an irregularly shaped pupil \cite{Aniridia})}.
\end{description}

For both scenarios the ISO/IEC standard specifies medical conditions that may apply. The first case includes excessive dilation or constriction of the pupil (associated with disease, trauma or abuse of drugs and alcohol), illnesses that affect the iris itself and the cornea, behind which the iris is located (\eg iritis, micro- and megalocornea, keratitis, leukoma), the aforementioned congenital diseases, such as aniridia or iris hypoplasia (underdevelopment), surgical procedures (cataract surgery, laserotherapy in glaucoma treatment, iridectomy) along many other pathologies -- disease, age or injury-related.

It is also possible to imagine a third scenario, that is not explicitly defined by the ISO/IEC standard:

\begin{description}[Scenario 3]
	\item [\textbf{Scenario 3}]{in which patient registered in a biometric system while already suffering from certain eye disease undergoes a treatment (\eg a lens replacement surgery in cataract conditions); this may lead to worse recognition when compared to condition before the treatment}.
\end{description}

\begin{svgraybox}
	Although ISO/IEC standard focuses on evaluating the impact of certain eye disorders on iris recognition, it is extremely difficult to gather data that represents diseases that are isolated from others, as different eye disorders often occur simultaneously. Therefore, a different approach to classifying disease influence may be proposed. This incorporates assembling selected medical conditions into groups that represent certain types of impact on the eye structures, and thus may affect iris recognition performance in different ways. These include eyes with no visible change, but with disease present, eyes with geometrically distorted pupils and irises, eyes with irises altered or damaged or, finally,  eyes with irises and pupils obstructed by pathological objects located in front of them. Such classification approach can stimulate more practical conclusions, such as putting forward changes in the eye that lead to worse recognition, which would be easy to identify during visual inspection by an expert.
\end{svgraybox}

\subsection{Terms and definitions}

\begin{description}[Terms and formulas]
\item [\textbf{FNMR}]{\textbf{False Non-Match Rate} -- an estimator of the false non-match probability, defined as considering a sample as not matching a given template when only samples matching this template are presented (\emph{i.e. genuine} attempts -- presenting same-eye iris images); FNMR is a function of an acceptance threshold $\tau$:
$$FNMR(\tau)=\frac{\#~of~attempts~not~matching~the~template(\tau)}{\# of all~genuine~attempts}$$}

\item [\textbf{FMR}]{\textbf{False Match Rate} -- an estimator of the false match probability, defined as considering a sample as matching a given template while only samples not matching this certain template are presented (\emph{i.e. impostor} attempts -- presenting iris images of different eyes); FMR is a function of an acceptance threshold $\tau$:
$$FMR(\tau)=\frac{\#~of~attempts~matching~the~template}{\#~of~all~impostor~attempts}$$}

\item [\textbf{EER}]{\textbf{Equal Error Rate} -- an acceptance threshold $\tau$ value, at which both False Match Rate and False Non-Match Rate are equal:
$$EER=FMR(\tau)=FNMR(\tau)$$}

\item [\textbf{FTE}]{\textbf{Failure To Enroll} -- a proportion of samples that could not be enrolled to the overall number of samples, from which there was an enrollment attempt:
$$FTE=\frac{\#~of~failed~attempts~to~create~a~template}{\#~of~all~attempts~to~create~a~template}$$}
\end{description}

\section{Description of medical conditions}
\label{sec:Medical}
\subsection{Cataract}
\begin{figure}[!h]
\centering
  \includegraphics[width=0.49\linewidth]{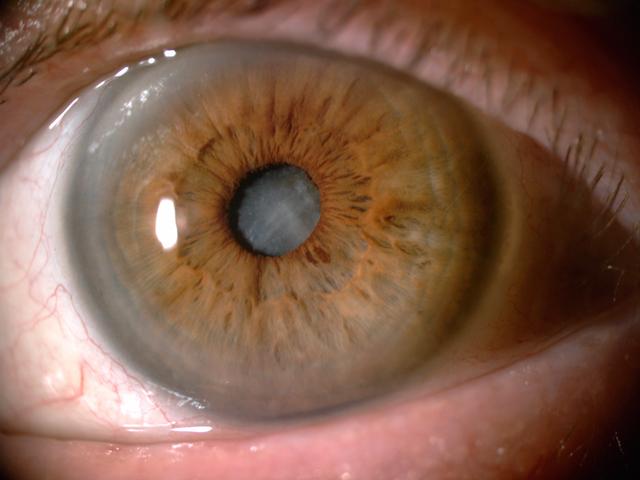}
  \includegraphics[width=0.49\linewidth]{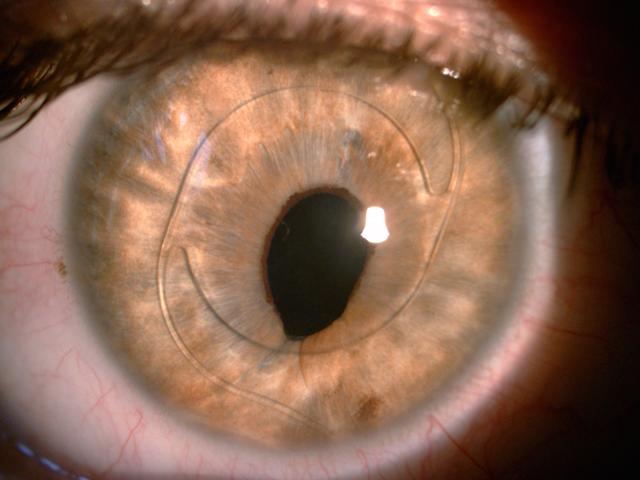}
\caption{\textbf{Left:} cataract-induced lens clouding. \textbf{Right:} artificial lens placed in the anterior chamber instead of behind the iris.}
\label{cataract}
\end{figure}

\runinhead{Disease characteristics.}
Cataract is one of the most common eye pathologies, being a complete or partial lens opacification leading to loss of transparency and ability to properly focus light onto the retina, Fig. \ref{cataract}. This results in a blurred and dimmed vision. Cataract accounts for as much as 30\% of blindness and visual impairment worldwide, mostly in developing countries. 

Cataracts may be classified twofold: congenital cataracts, that occur even before the birth or developing in the first few years of life, caused by congenital metabolic disorders, ionizing radiation, certain drugs (sulfonamides, corticosteroids), infections in the uterus. Th second type are the more common, usually age-related cataracts or secondary cataracts, induced by systemic diseases, accompanying other eye pathologies, or as a result of trauma, ionizing radiation, or toxins \cite{Nizankowska}.

\runinhead{Treatment incorporating lens replacement.}
Cataract treatment focuses solely on a surgical procedure of lens extraction (called \emph{phacoemulsification}) using aspiration, during which a small probe is inserted through an incision in the side of the cornea and then through a circular hole in the lens capsule to reach the lens. The probe then emits ultrasound waves to break the opacified lens, which is later removed using suction. The resulting condition, \emph{aphakia}, can be compensated using either prescription glasses, contact lenses, or an intraocular implant (artificial lens) with individually calculated focusing power \cite{TrokielewiczWilga2014}.

\runinhead{Influence on the remaining eye structures.}
Cataract itself should not have any significant impact on other parts of the eyeball, including the iris. However, there are many other pathologies that accompany cataract and may cause damage or alteration to the iris \cite{TrokielewiczWilga2014}, for instance:
\begin{itemize}
\item acute glaucoma (possible flattening of the iris and pupil distortion),
\item anterior and posterior synechiae (iris adhered to either the lens or to the cornea, respectively, Fig. \ref{synechiae_cataract}),
\item rubeosis iridis (pathological vascularization in the iris),
\item iris atrophy,
\item pseudoexfoliation syndrome (accumulation of protein fibers inside the eye),
\item conditions after iris laserotherapy,
\item post-traumatic cataract.
\end{itemize}

In addition to this, in some cases the surgical cataract extraction incorporates placing the lens implant not behind, but in front of the iris. This happens when there is not enough capsular support remaining after the lens extraction and in such cases the implant has to be attached to the circumferential part of the iris or supported on the iridocorneal angle using claw-shaped hooks, Fig. \ref{cataract}. This may affect the look of the iris and even distort the circular shape of the pupil \cite{TrokielewiczWilga2014}.

\subsection{Glaucoma}
\runinhead{Disease characteristics.}
Glaucoma is a group of ocular disorders, typically described as damage to the optical nerve followed by visual impairment as a result of increased intraocular pressure, however, in certain cases these may happen with low or normal pressure levels. Therefore, glaucoma should be given a description of a multi-factorial optical nerve neuropathy, with increased intraocular pressure being a risk factor \cite{Nizankowska}. Glaucoma is a second-leading cause of visual impairment and blindness worldwide, secondary to cataracts only \cite{GlaucomaWHO}.

\begin{figure}[h!]
\centering
  \includegraphics[width=0.49\linewidth]{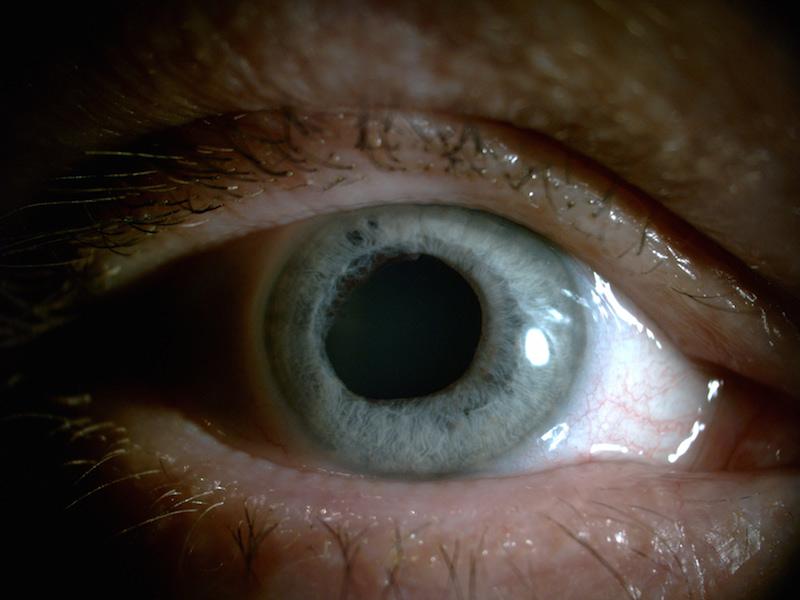}
    \includegraphics[width=0.49\linewidth]{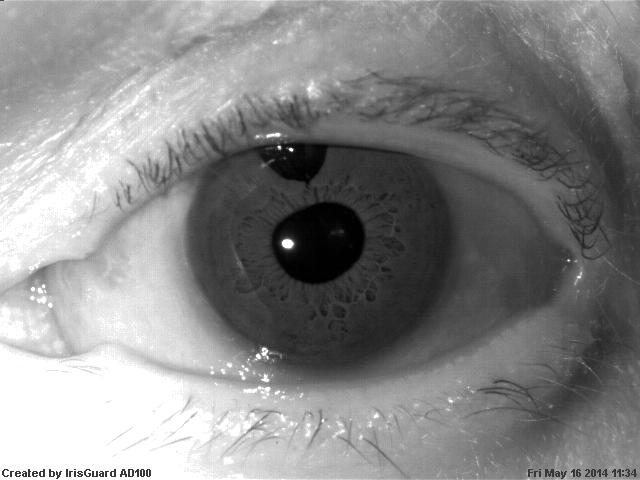}
\caption{\textbf{Left:} a distortion in pupil shape resulting from acute (closed-angle) glaucoma. \textbf{Right:} iris tissue damage due to the iridectomy procedure.}
\label{glaucoma}
\end{figure}

\runinhead{Glaucoma and its impact on iris.}
This disease can be subdivided into two main categories, namely the open-angle and the closed-angle glaucoma. The former type is chronic and often does not cause any pain as it progresses slower than the latter. The second type of glaucoma, known as the acute glaucoma, usually happens suddenly when the angle between the iris and the cornea closes completely, preventing the aqueous humor from flowing through the trabecular meshwork towards the inside of the eyeball, causing a sudden increase of the intraocular pressure and severe pain to the patient. Aqueous humor pushing against the iris and the lens may cause certain flattening of the iris and distort the shape of the pupil, as shown in Fig. \ref{glaucoma}. While the first symptom is often not visible, the other one is significant and can contribute to a decrease in iris recognition performance as it affects the segmentation procedures that often approximate iris and pupil boundaries with circles. Also, an increase in the intraocular pressure can cause a corneal edema reducing the cornea's transparency, and thus making it difficult to get a clear image of the iris behing it. 

\runinhead{Treatment.}
Glaucoma treatment, usually associated with reducing the pressure inside the eyeball, incorporates either a surgically performed, triangle-shaped incision in the upper part of the iris (iridectomy) or making a small, circular puncture also in the upper part of the iris (iridotomy). Obviously, both these procedures affect the look of the iris. However, the incision created during iridotomy is usually very small and hidden under the upper eyelid, and thus rarely visible. The triangle-shaped cutout, however, is rather large and may significantly affect the look of the iris, Fig. \ref{glaucoma}. 

\subsection{Posterior and anterior synechiae}
\begin{figure}[h]
\centering
  \includegraphics[width=0.49\linewidth]{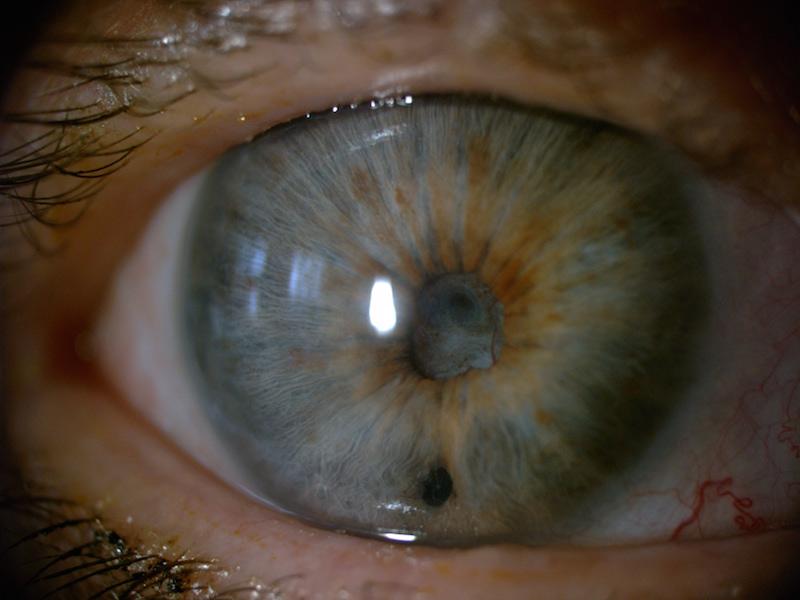}
  \includegraphics[width=0.49\linewidth]{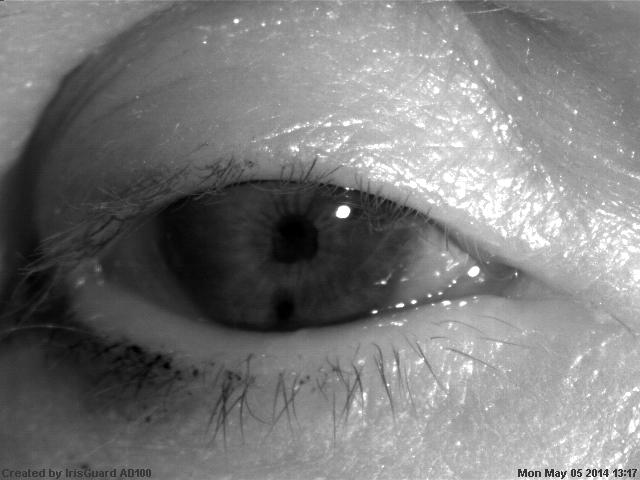}
\caption{\textbf{Left:} posterior synechiae seen in visible light. \textbf{Right:} same, but in NIR illumination. The pupil appears dark in the NIR image, in contrast to the visible light image.}
\label{synechiae}
\end{figure}

\begin{figure}[!h]
\centering
  \includegraphics[width=0.49\linewidth]{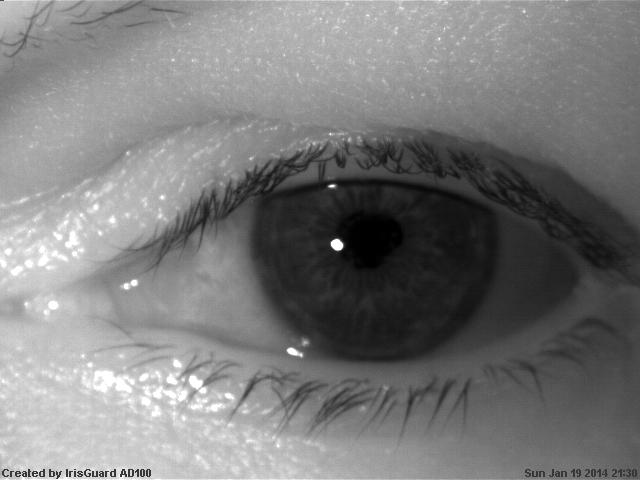}
  \includegraphics[width=0.49\linewidth]{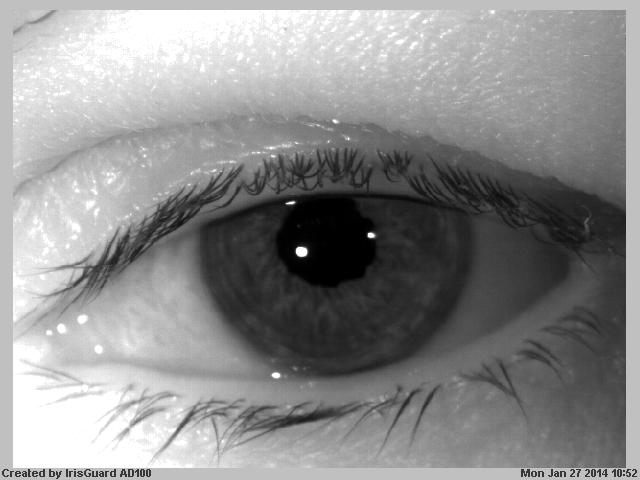}
\caption{Cataract eye with posterior synechiae before ({\bf left}) and after the procedure of lens removal ({\bf right}). Distortion of the inner iris boundary remains even after the lens has been extracted and the synechiae have been removed.}
\label{synechiae_cataract}
\end{figure}

\textbf{Synechiae} occur when the iris becomes attached either to the lens (posterior synechiae) or to the cornea (anterior synechiae). They may be caused by various conditions, in most cases by ocular trauma, iritis (inflammation of the iris) or other forms of uveitis (inflammation of the uvea). \textbf{Posterior synechiae} become clearly visible when pharmacological pupil dilation is performed and can alter the shape of the pupil significantly, as well as bring opacification to the surface of the lens, making it brighter than normally, Fig. \ref{synechiae}. When synechiae accompany cataract, after the lens extraction there may be deformation left, even though the iris is no longer attached to the lens, Fig. \ref{synechiae_cataract}. In some cases, this condition may lead to glaucoma, as the aqueous humor flow is partially blocked. \textbf{Anterior synechiae}, when the iris adheres to the cornea, often lead to glaucoma due to closing the angle between the iris and the cornea and blocking the flow of the aqueous humor through the trabecular meshwork. This condition, however, does not directly affect the look of the iris or the pupil.

\subsection{Retinal detachment}
\begin{figure}[h]
\centering
\includegraphics[width=0.49\linewidth]{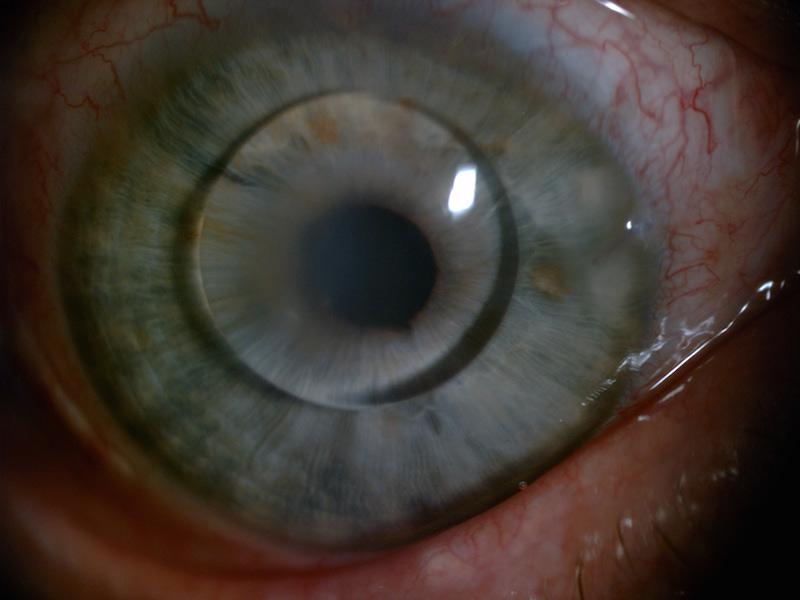}
\includegraphics[width=0.49\linewidth]{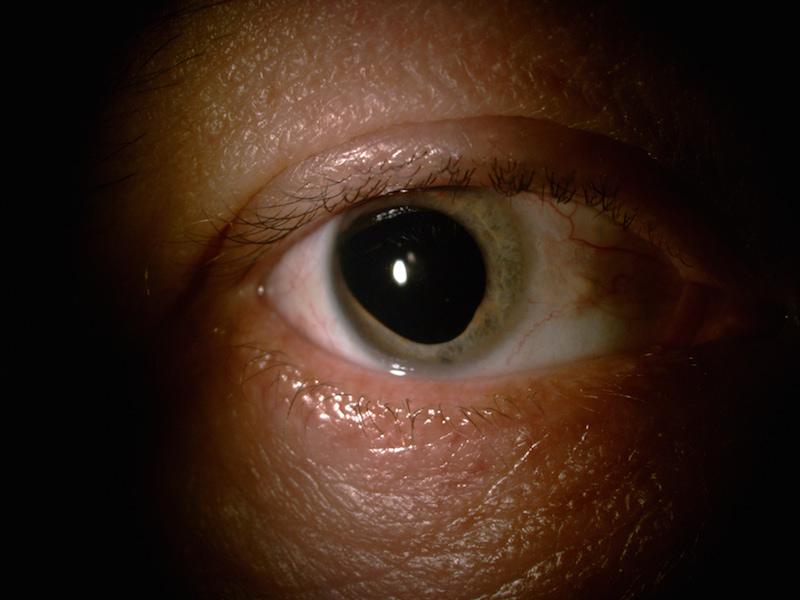}
\caption{Conditions that may accompany retinal detachment or occur during treatment: silicon oil in the anterior chamber of the eye ({\bf left}), distortion of the pupil ({\bf right}).}
\label{retina}
\end{figure}

Subdividing into many conditions, \textbf{retinal detachment} can be roughly summarized as a disorder, in which the retina -- a light-sensitive part of the internal membrane of the eyeball -- detaches from the underlying layers, creating a severe visual impairment leading to complete blindness if not treated immediately \cite{Nizankowska}. The illness itself does not have a direct impact on the iris or on the lens. The treatment, however, incorporates filling the eyeball with silicon oil to push the detached retina towards the back of the eyeball and then fusing it in place using laser photocoagulation. The oil is then removed and replaced with artificial fluid that substitutes the vitreous humor. In certain cases, the oil may flow back to the anterior chamber of the eye, Fig. \ref{retina}, left. This creates an obstruction that alters the look of the iris. We also came across one case in which the retinal detachment was accompanied with wide, oval, non-reacting pupil with partial iris atrophy, Fig. \ref{retina}, right. These are obvious candidates to cause potential trouble for iris recognition.

\subsection{Rubeosis iridis}
\begin{figure}[!h]
\centering
  \includegraphics[width=0.49\linewidth]{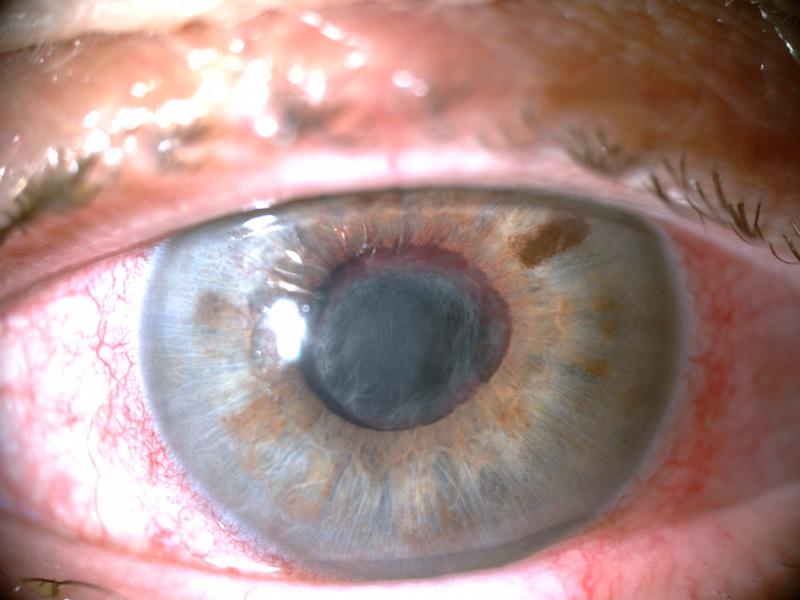}
  \includegraphics[width=0.49\linewidth]{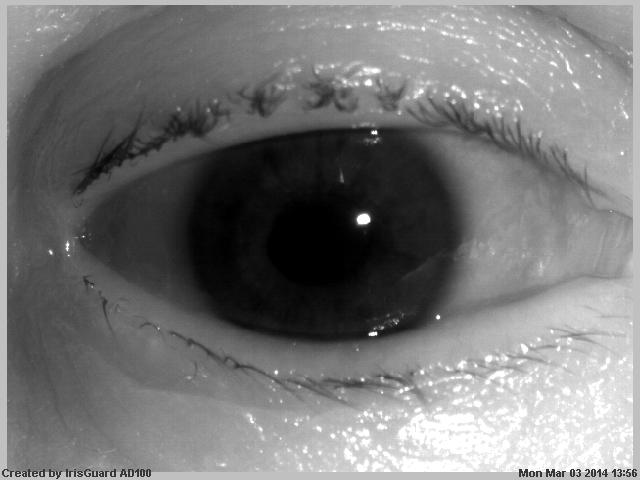}
\caption{Pathological angiogenesis in the iris tissue -- rubeosis iridis. Visible light ({\bf left}) and NIR ({\bf right}) illumination images show differences in the appearance of iris tissue populated with blood veins. The NIR image is much less influenced by this type of pathology.}
\label{rubeosis}
\end{figure}

\textbf{Rubeosis} is a pathological vascularization process afflicting the surface of the iris, often as a result of disease present in the retina, Fig. \ref{rubeosis}. Ischemic retina can release vessel growth factors to meet its oxygen needs, however, such vascularization is not a normal condition and may clog the iridocorneal angle, causing an increase in intraocular pressure \cite{Rubeosis}. This alters the look of the iris surface with the potential to affect iris recognition. However, it is worth noticing that NIR light transmission through blood is different than that of visible light, therefore the NIR-illuminated images may show this pathology as far less obtrusive than the visible light photographs. See Fig. \ref{rubeosis} for comparison of images taken with those two types of illumination.

\subsection{Other eye pathologies}
\begin{figure}[htbp]
  \centering
  \includegraphics[width=0.49\linewidth]{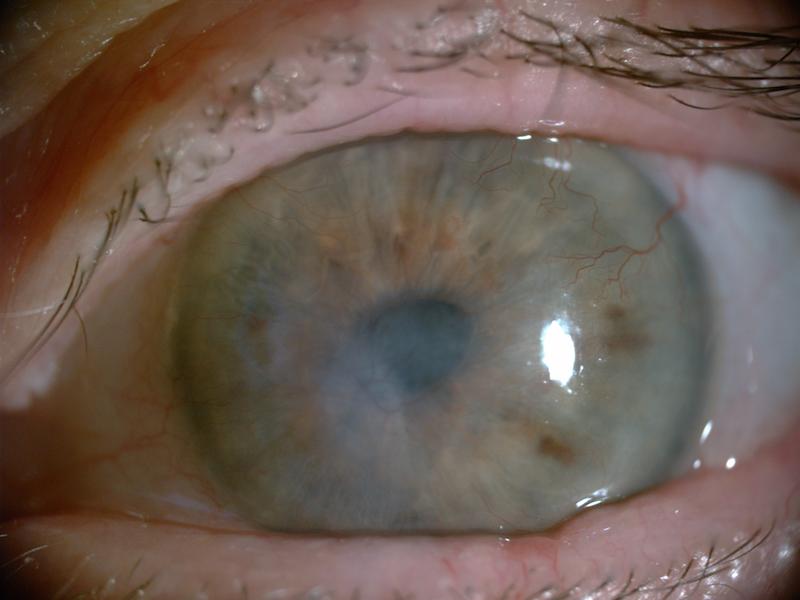}
  \includegraphics[width=0.49\linewidth]{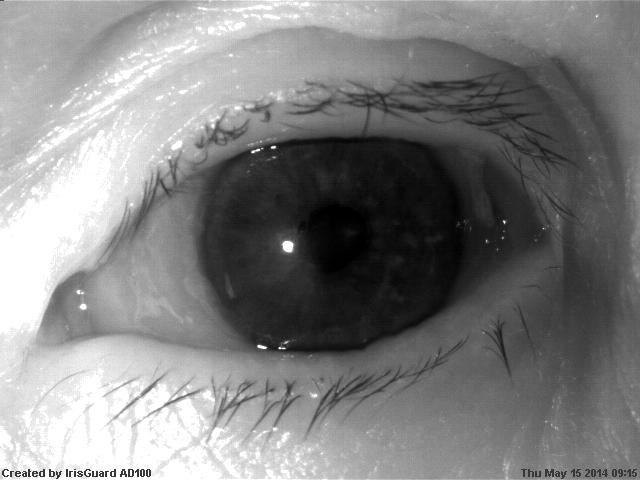}
\caption{Pathological angiogenesis in the cornea creates a haze obstructing the view of the iris below it. Visible ({\bf left}) and NIR ({\bf right}) illumination comparison shows much less impact of this type of pathology with the latter.}
\label{corneal_vasc}
\end{figure}

\begin{figure}[htb!]
\sidecaption[t]
\includegraphics[width=0.6\linewidth]{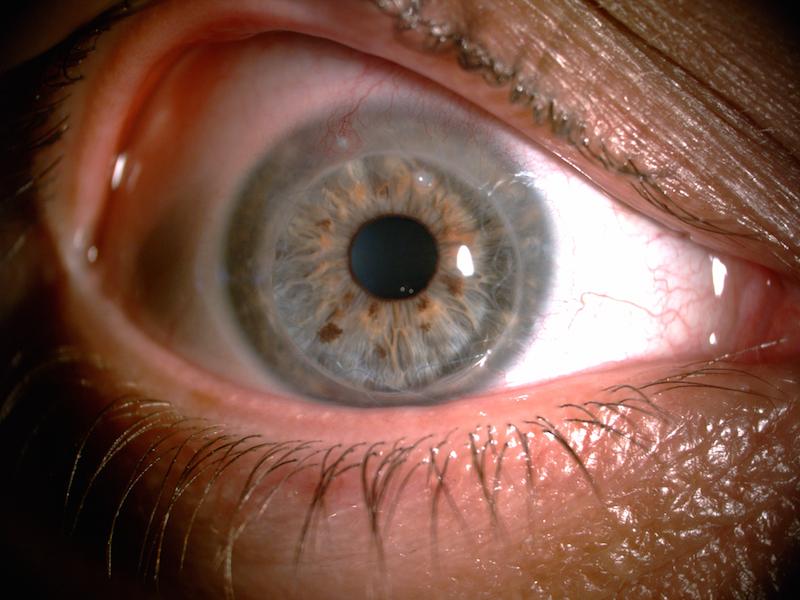}
\caption{Grafted cornea with visible sutures obstructing the view of the iris. Also, the newly implanted cornea is far more transparent than the remainings of the original tissue.}
\label{corneal_graft}  
\end{figure}

During the process of data collection, we came across several less common eye pathologies, illnesses or conditions, often accompanying other disorders, but sometimes occurring independently. Those are divided in respect to the part of the eye they influence most. Disorders affecting the cornea were: 
\begin{itemize}
\item pathological \textbf{vascularization (angiogenesis)} in the corneal tissue, causing occlusions that partially prevent the light from entering the eye, and also obstructing the view of iris pattern, Fig. \ref{corneal_vasc},
\item \textbf{corneal haze, ulcers or opacities} of different origin, with consequences similar to the ones associated with pathological angiogenesis (obstruction of the iris pattern),
\item \textbf{corneal grafting} with grafts sutured and therefore obstructing the view of the iris, Fig. \ref{corneal_graft}.
\end{itemize}

\begin{figure}[htb!]
\sidecaption[t]
\includegraphics[width=0.6\linewidth]{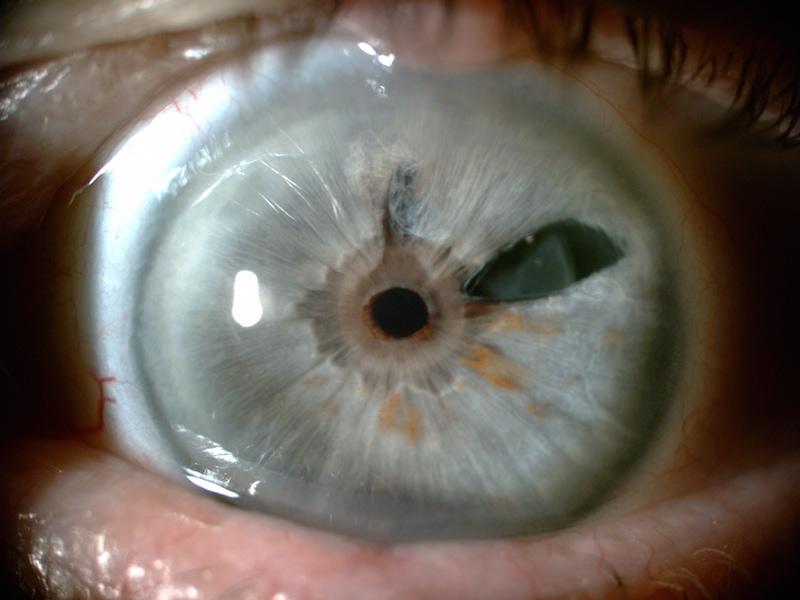}
\caption{Rupture-like damage to the iris tissue revealing the lens underneath it and altering the iris pattern by removing significant portions of the iris tissue.}
\label{iris_damage}  
\end{figure}

Disorders affecting the iris were:
\begin{itemize}
\item \textbf{sutures} in the iris,
\item \textbf{iris dialysis} with a piece of iris tissue detached from the base of the iris,
\item \textbf{damage or atrophy to the iris tissue} resulting in missing fragments of its structure, Fig. \ref{iris_damage}.
\end{itemize}

\section{Related work} 
\label{sec:Related}
\subsection{Cataract and phacoemulsification procedure influence}
A study by Roizenblatt \etal \cite{Roizenblatt} involved 55 patients suffering from cataract. Enrollment for each eye was performed using an iris biometric system (LG IrisAccess 2000) before cataract extraction. Then, verification was performed, three times before the surgery, and three times after the cataract extraction (the latter three verification trials were performed 30 days after the procedure and 7 days after stopping the administration of pupil-dilating drugs, \ie a time period, after which healing should be finished). After such time period, pupils are expected to revert to their normal size and reaction to light, which is later confirmed by authors' observations, revealing differences in the size no larger than 1.5mm when compared with images collected before the treatment. The verification experiments showed an increase in Hamming distance when average HD score obtained in post-surgery trials (HD=0.2094) is compared with average HD score obtained in pre-surgery trials (HD=0.098). The biometric system failed to recognize 6 out of 55 eyes (FNMR of approx. 11\% is reported). It is worth noticing that authors used a very liberal acceptance threshold of 0.4. To come up with possible explanation of worse performance, the authors assigned a visual score between 0 and 4, given by an ophthalmology surgeon, to each of the eyes that underwent the cataract extraction procedure. One point was assigned for each of the following ocular pathologies: depigmentation, pupil ovalization, focal atrophy with and without transillumination. Statistical analysis revealed a correlation between iris pattern deviation intensity and HD shift towards worse (\ie higher) scores. Authors suggest that probe manipulation and energy released in the eyeball during the procedure may cause atrophic changes to the iris tissue. They also make a claim to be able to predict trouble with iris recognition based on a visual inspection of the eye that underwent the procedure. In such cases, re-enrollment is suggested.  

Phacoemulsification refers to the extraction of the lens through aspiration. The procedure involves the insertion of a small probe through an incision in the side of the cornea. The probe emits ultrasound waves that break the opaque lens which is later removed using aspiration \cite{TrokielewiczWilga2014}. Seyeddain \etal \cite{Seyeddain2014} conducted a study on the effects of phacoemulsification and pharmacologically induced mydriasis on iris recognition. The experiment aimed to determine whether irises, following phacoemulsification or drug induced mydriasis (preventing the dilated pupil from reacting to light stimulation) perform worse when compared to the same irises before the procedure or before the drug-induced pupil dilation. They revealed that 5.2\% of the eyes subject to cataract surgery could no longer be recognized after the procedure. In the pupil dilation group, this portion reached as high as 11.9\%. In both cases the authors suggest re-enrollment for patients whose eyes were not successfully identified after the surgery or instillation of mydriatics. No false acceptances were observed in either case.

In our previous work \cite{TrokielewiczWilga2014} we presented an experimental study revealing weaker performance of the automatic iris recognition methods for cataract-affected eyes when compared to healthy eyes. A database of 1288 eye images coming from 37 ophthalmology patients patients has been gathered. To assess the extent of recognition accuracy deterioration, three commercial and academic iris recognition methods were employed to calculate genuine match scores for healthy eyes and those with cataracts. A significant degradation in recognition accuracy was shown for all three matchers used in this study (12\% of genuine score increase for an academic matcher, up to 175\% of genuine score increase obtained for an example commercial matcher). False-non match rates were affected by cataract-induced changes in two out of three iris matchers. 

In a study by Dhir \emph{et al.} \cite{Dhir}, the effects of the use of mydriatics accompanying cataract surgery, as well as the effects of the procedure itself, are examined. A group of 15 patients had their eyes enrolled prior to cataract extraction surgery. Four verification attempts were then performed: 5, 10, and 15 minutes after the application of pupil-dilating drugs (but still before the surgery), and finally -- 14 days after the surgery. Pupil dilation due to mydriatics use caused Hamming distances to gradually increase as time after drug instillation elapsed. This led to FNMR of 13.3\% (6 out of 45 verification attempts failed). Surprisingly, none of the eyes were falsely rejected in the verification attempt conducted 14 days after the cataract removal procedure. However, the authors excluded from the dataset eyes with pre-existent corneal and iris pathologies, or those with iris tissue damaged during the surgery. This brings a certain bias to the analysis, thus the results may not reflect a real-world application.  The study suggests that decreased recognition accuracy after the cataract surgery originates from a slight shift of the iris towards the center of the eyeball as a result of implanting an artificial lens that is thinner than the natural one. Specular reflections from the implant may also contribute when certain localization methods are employed. Authors also warn that excessive pupil dilation can be exploited by criminals in order to enroll under multiple identities to deceive law enforcement.

Recently, Ramachandra \etal \cite{RamachandraCataractICB2016} conducted experiments regarding iris biometrics in the context of cataract surgery, when verification is performed using pre-surgery gallery samples and post-surgery probe samples, coming from the same individual. Studies were carried out using a database of iris images acquired from 24 hours pre-surgery and 36-42 hours post-surgery from 84 subjects. Recognition accuracy is reported to drop significantly, reaching genuine match rate of 85.19\% @ FMR=0.1 and EER=7\%, when compared to performance achieved using pre-surgery images only.

\subsection{Refractive surgeries}
Laser-assisted refractive correction surgeries and their possible impact on the accuracy of iris biometrics is studied by Yuan \etal \cite{Yuan}. These procedures involve making an incision in the cornea to create a flap with a hinge left on one side. The flap is then lifted to expose the middle part of the cornea, which is then ablated using short pulses of a 193 nm excimer laser to achieve a finely-tuned shape that depends on the treated condition. The corneal flap is then folded back and eye is left to heal itself. These methods are commonly known as \emph{Laser-Assisted in-Situ Keratomileusis} method, abbreviated \emph{LASIK}. \emph{LASIK} procedures are widely used across the globe to improve life quality for patients suffering from myopia, hypermetropia, or astigmatism. An experiment is thus carried out to find out whether such manipulation may result in false non-matches when iris biometrics is employed. Using Masek's method, 13 eyes out of 14 were correctly recognized after the procedure. However, the one eye that was falsely rejected had a significant deviation in circularity of the pupil and increased pupil diameter. The authors argue that refractive correction procedures have little effect on iris recognition. Nonetheless, more experiments involving larger datasets are called for. 

\subsection{Other ocular pathologies}
Aslam \emph{et al.} \cite{Aslam} conducted one of the most extensive studies in terms of disease representation in the database. 54 patients suffering from various diseases had their eyes imaged with an IrisGuard H100 sensor and then enrolled in a biometric system based on Daugman's idea, before they underwent any treatment. Then, verification attempts were carried our after the treatment and Hamming distances (HD) between iris codes obtained before and after the treatment were calculated to tell whether medical procedures applied had any impact on recognition accuracy. Iris recognition method employed by the authors turned out to be resilient for most illnesses, \emph{i.e.}, glaucoma treated using laser iridotomy, infective and non-infective corneal pathologies, episcleritis, scleritis and conjunctivitis. However, 5 out of 24 irises affected by anterior uveitis, a condition in which the middle layer of the eye, the uvea, which includes the iris and the ciliary body, becomes inflamed \cite{Uveitis}, were falsely rejected after the treatment, hence FNMR=21\% in this particular subset. Rejected eyes had earlier been administered with mydriatics and thus had pupils significantly dilated. In addition, two eyes suffered from high corneal and anterior chamber activity, while the remaining three had posterior synechiae that caused deviation from the pupil circularity. Hypothesis stating that the mean HD in the anterior uveitis subset is equal to mean HD in the control group consisting of healthy eyes has been rejected with $p < 10^{-4}$, while there were no statistically significant differences between scores obtained from other disease subsets when compared to the control group. For pathologies related to corneal opacities, Aslam tries to explain the lack of recognition performance deterioration by the fact that NIR illumination used in iris biometrics is more easily transmitted through such objects and therefore allows correct imaging of underlying iris details. Laser iridotomy also showed little influence, as the puncture in the iris tissue made by laser appears to be too small to alter the iris pattern significantly. However, certain combinations of synechiae and pupil dilation can affect the look of the iris texture vastly enough to produce recognition errors. A deviation in pupil's circularity caused by synechiae may also contribute to segmentation errors.

Borgen \emph{et al.} \cite{Borgen} conducted a study focusing on iris and retinal biometrics in which they use 17 images chosen from the UBIRIS database and then digitally modified to resemble changes to the eye structures caused by various ocular illnesses: keratitis and corneal infiltrates, blurring and dulling of the cornea, corneal scarring and surgery, angiogenesis, tumors and melanoma. High FNMR values (32.8\% -- 86.8\%) are reported for all modifications, except for the pathological vascularization (6.6\%), changes in iris color (0.5\%) and iridectomy, for which FNMR=0\%. Faulty segmentation is supposed to be the main cause, especially in cases with present corneal clouding. Authors, however, do not acknowledge the fact that near-infrared illumination enables correct imaging even in eyes with corneal pathologies such as clouding or other selected illness-related occlusions. 

McConnon \emph{et al.} \cite{McConnon2012} studied disorders causing pupil/iris deformation, pupil/iris occlusion and those with no iris or with a very small iris, to estimate the impact they may have on the reliability of iris image segmentation.  Encoding and matching are not executed in their study. However, due to lack of publicly available datasets, the authors are forced to use images coming from the Atlas of Ophthalmology. These are photographs obtained in visible light, and thus not always suited for iris recognition, which typically uses NIR-illuminated, 8-bit grayscale images. The dataset was downsampled to $320\times240$ resolution and manually segmented to obtain the ground truth iris localization. The authors then performed the automatic segmentation using Masek's algorithm to find out if the results would vary from those obtained when segmenting the images manually. The results suggest that segmentation stage can be influenced by the presence of the foretold pathologies, as automatic segmentation deviated, when compared to manual segmentation, by two or more pixels in 46\% and 55\% of images for the limbic and pupillary boundaries, respectively.

Our most recent work devoted to this subject expands earlier experiments regarding cataract-related effects and brings the novelty of assessing which types of eye-afflicted damage caused by diseases have the greatest impact on the accuracy of biometric systems employing iris recognition. Changes to the iris tissue and geometrical distortions in the pupillary area are shown to have the highest chance of degrading genuine comparison scores for three different iris recognition algorithms employed in that research \cite{TrokielewiczCYBCONF2015}. This study was later extended with more data collected from ophthalmology patients over a longer period of time, as well as with experiments that revealed an increased chance of failure-to-enroll errors when iris biometric systems are presented with images obtained from patients with diseased eyes \cite{TrokielewiczBTAS2015}. The samples that were used for these experiments were selected so that they comply with the ISO/IEC 29794-6 standard for iris image quality. Segmentation errors were suggested as the most probable source of deteriorated matching accuracy. The datasets of eye images affected by ocular disorders, used in the preliminary and the expanded research, are publicly available to all interested researchers \cite{WarsawDiseaseIris1, WarsawDiseaseIris2}. Further extension of this research, including experiments conducted with the use of an additional iris recognition method can be found in \cite{TrokielewiczDiseasesIMAVIS}. 

\subsection{Post-mortem iris recognition}
Post-mortem iris recognition in humans can certainly be classified as the most extreme case of 'ocular pathology'. To establish whether iris can still serve as a biometric identifier after death is crucial from the forensics point of view: \emph{`Can it provide a tool for forensics examiners in cases when other biometric methods may be unavailable or too burdensome?'}, as well as because of security issues: \emph{`Can the iris be stolen after death and used in presentation attacks?'}.

The authors' studies regarding this important, yet maybe unpleasant topic, revealed that images obtained 27 hours post-mortem can still be successfully recognized when matched against samples obtained shortly after demise (with accuracy reaching 70\%), despite common statements about iris recognition being impossible to employ after death \cite{TrokielewiczPostMortemICB2016}. This study was later expanded to examine whether this can still be considered true when images obtained after even longer periods of time are considered. Paper \cite{TrokielewiczPostMortemBTAS2016} extends the earlier work with images obtained up to 17 days post-mortem, revealing that iris recognition can still occasionally work even as long as 407 hours after a person's death. Moreover, the dataset collected for the purpose of these studies is made publicly available to all interested researchers to facilitate research in this field \cite{WarsawColdIris1}. Other reports on post-mortem iris recognition can be found in Saripalle \etal \cite{PostMortemPigs} and Sansola \cite{BostonPostMortem}. 

\section{Database of iris images collected from patients with ocular disorders}
\label{sec:Database}
We had a rare opportunity of close cooperation with an ophthalmologist's office, which provided us with the dataset that represents various medical conditions affecting the iris and its surrounding structures. This database is detailed in the following section. 

\subsection{Data collection process}
The data collection process was carried out during approximately 16 months of typical patients' visits to the ophthalmologist's office. During its course, all patients attending visits had their eyes photographed with a professional iris recognition camera operating in near-infrared spectrum (IrisGuard AD100) and, in selected cases, also with two cameras operating in visible spectrum (Canon EOS 1000D with EF-S 18-55 mm f/3.5-5.6 lens equipped with a Raynox DCR-250 macro converter and a ring flashlight for macrophotography, and an ophthalmology slit-lamp camera Topcon DC3). The two latter cameras are employed in particularly interesting cases to perform visual inspection of the illnesses' impact in samples showing significant changes to the eye structures. 

During the first visit, a typical ophthalmology examination was performed by an ophthalmologist, and the patient was enrolled anonymously into the system. Then, at least six photographs of each eye have been captured. Images within a single acquisition session were acquired in separate presentations, as recommended by ISO/IEC 19795-2, \emph{i.e.}, the patient was asked to lift his/her head from the chin and forehead rests after each capture. This is to purposefully introduce some noise in the intra-session sample sets. This procedure has been then repeated with future visits, each of which had a separate metadata, \emph{e.g.}, to distinguish between pre- and post-surgery examinations when such treatment is applicable and has been performed. 

\subsection{Database description and statistics}
The entire dataset comprises 2996 images collected from 230 distinct irises. Each class, corresponding to each iris, is represented by NIR images collected with the IrisGuard camera, while some classes are also represented by color images. Regarding session count, for 184 image classes there are samples collected in one acquisition session. 38 classes contain images collected in two sessions, 6 classes contain images collected in three sessions, and finally for 2 classes there are four different image acquisition sessions. The second and subsequent sessions are usually conducted after some kind of medical procedures or treatment (a cataract extraction surgery, for instance). 

\subsection{Access to the database}
Papers \cite{TrokielewiczCYBCONF2015} and \cite{TrokielewiczBTAS2015} present the described dataset as a package publicly available to interested researchers in version 1 and an expanded version 2, respectively. The released database also contains an extensive medical description for each eye represented in the data, including all diseases present in the eye, medical procedures, examinations, and other remarks.

\section{Experimental study and results}
\label{sec:Experiments}
\subsection{Data subsets creation methodology}
\runinhead{Criteria A: medical conditions.}
In the first part of this study we attempted to divide the data with respect to the type of illness or other medical condition present in the eye. Only one particular disorder is represented by enough data for a thorough analysis, thus in this section we discuss only the cataract eyes (before any surgical treatment) - referred to as \emph{Cataract} subset later on.

\runinhead{Criteria B: influence on different eye structures.}
Having a database that gathers various eye pathologies does not necessarily mean that each eye is affected by one illness only. In most cases, there are two or even more conditions present in one eye. Some of them do not affect the iris at all, some impact pupillary regions, other target the cornea or the iris tissue itself. This abundance of various and often unrelated medical conditions makes precise analysis of illness impact on iris recognition very difficult. Therefore, we come up with a solution of subdividing the dataset into several groups, connected not by the illnesses present in the eye themselves, but by the type of impact they have on iris or other structures of the eyeball, regardless of their medical origin. Such approach provides a possibility to analyze the impact of certain types of disorders on the recognition performance irrespective of an actual medical condition. It also makes it easier to propose countermeasures based on visual inspection of samples. 

\begin{figure}[htb]
\begin{center}
	\includegraphics[width=0.497\linewidth]{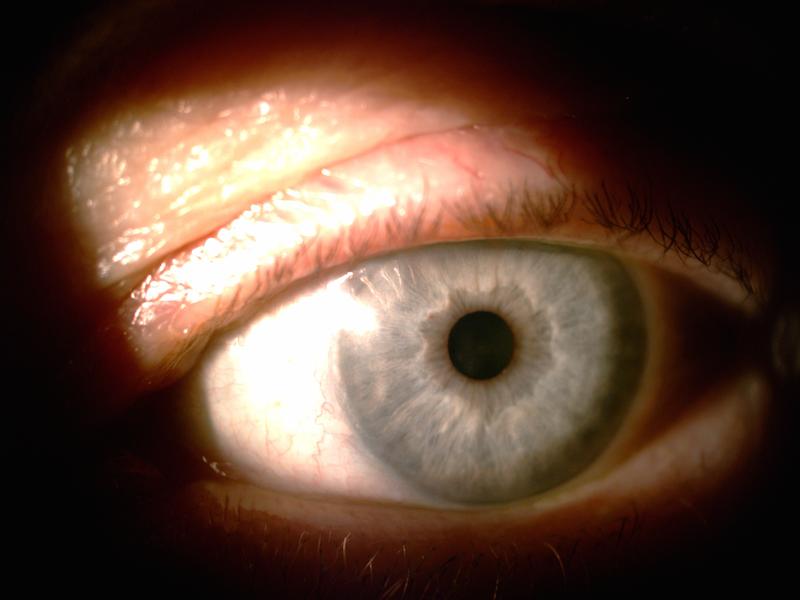}
	\hfill
	\includegraphics[width=0.497\linewidth]{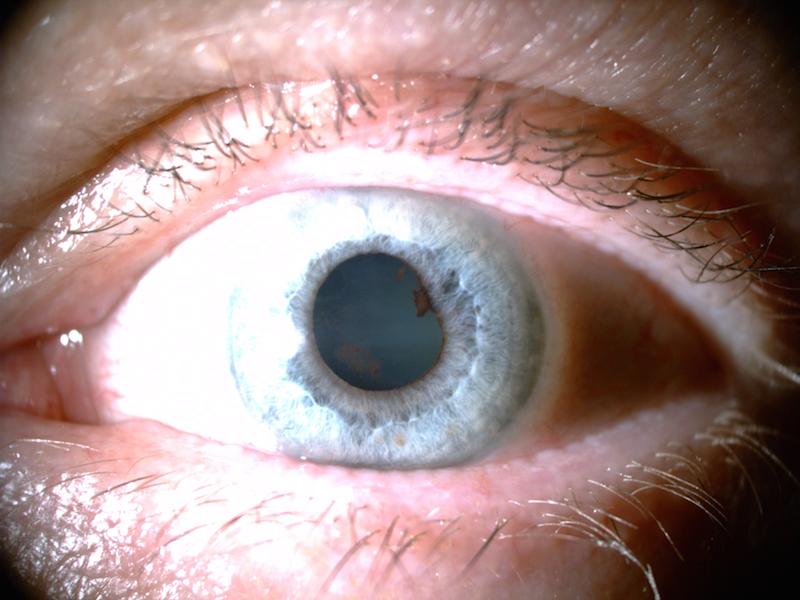}
	\vfill
	 \includegraphics[width=0.497\linewidth]{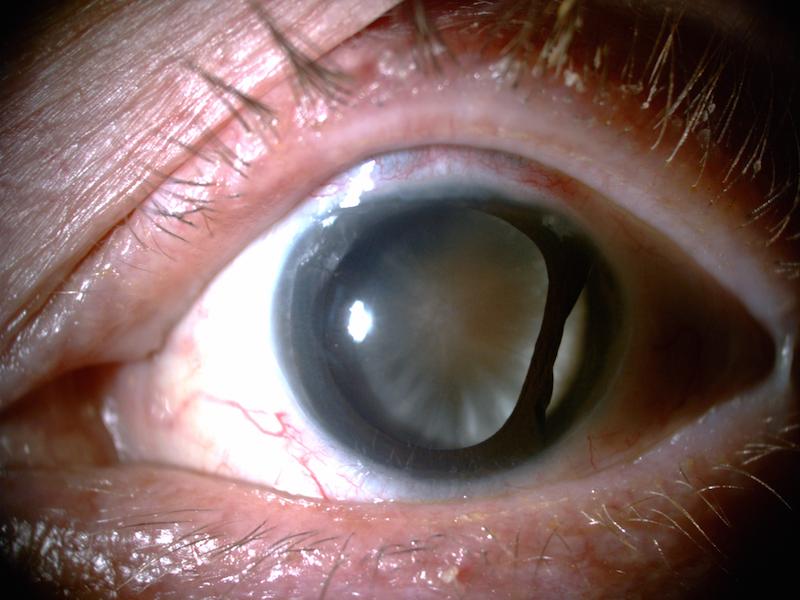}
  \hfill
  \includegraphics[width=0.497\linewidth]{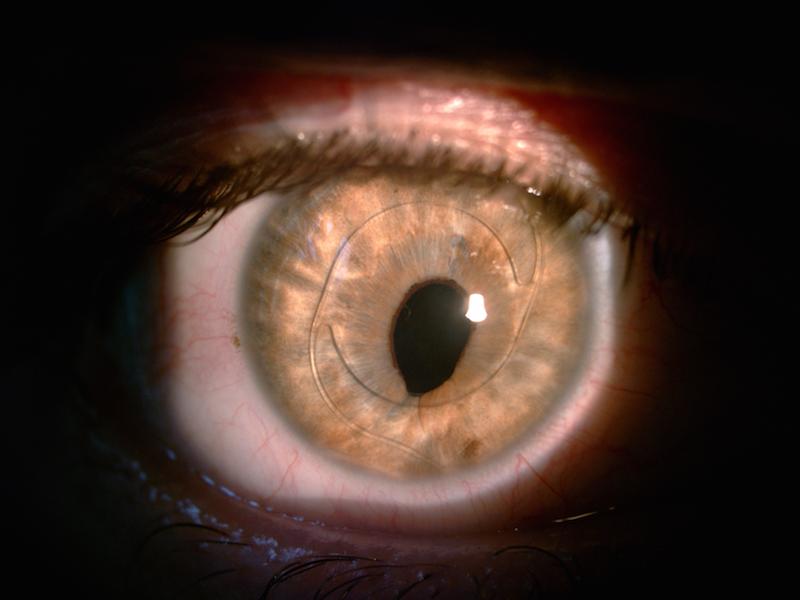}
\end{center}
\caption{Images obtained using Topcon DC3 slit-lamp microscope camera. \textbf{Top left}: a sample from the \emph{Clear} subset without any visible changes that may affect iris recognition. \textbf{Top right}: a sample from the \emph{Geometry} subset with posterior synechiae disturbing the shape of the pupil and the iris. \textbf{Bottom left}: a sample from the \emph{Tissue} subset with severe changes to the iris tissue. \textbf{Bottom right}: a sample from the \emph{Obstructions} subset lens implant placed in the anterior chamber of the eye, in front of the iris.}
\label{fig::topcon_samples}
\end{figure}

\begin{svgraybox}
Five major subsets can be distinguished on that basis: healthy eyes (referred to as \emph{Healthy} partition, serving as a control group), eyes with no visible changes (referred to as \emph{Clear} partition), eyes with changes in pupillary regions, such as deviation from pupil's circular shape (referred to as \emph{Geometry}), eyes with visible alterations to the iris tissue (referred to as \emph{Tissue}) and eyes with iris covered by obstructions located in front of it (referred to as \emph{Obstructions}), Fig. \ref{fig::topcon_samples}. Tab. \ref{database_summary_divided} provides detailed description and number of samples in each group. 
\end{svgraybox}

\begin{table*}
\renewcommand{\arraystretch}{1.1}
\caption{Database partitions created in respect to the type of illness impact on certain eye structures. Only NIR sample count is shown as these are used in experiments with automated iris recognition methods.}
\label{database_summary_divided}
\begin{tabular}{|p{0.35\textwidth}|p{0.40\textwidth}|p{0.10\textwidth}|p{0.10\textwidth}|}
\hline
\textbf{Influence type \emph{data partition}} & \textbf{Description} & \textbf{Number of eyes} & \textbf{Number of NIR samples}\\
\hline
\textbf{Healthy eyes} \emph{(Healthy)} & Eyes with no illness or pathology present & 35  &  216\\
\hline
\textbf{No visible changes} \emph{(Clear)} & Eyes with present medical condition, but not affected visually & 87  & 568 \\
\hline
\textbf{Changes in pupil/iris geometry} \emph{(Geometry)} & Distortions in pupil and iris circularity due to various medical conditions & 53 & 312\\
\hline
\textbf{Iris tissue alterations} \emph{(Tissue)} & Damage or atrophy of the iris tissue itself & 8 & 50\\
\hline
\textbf{Obstructed iris} \emph{(Obstructions)} & Objects or opacities in the eye structures located in front of the iris (\emph{i.e.}, cornea and anterior chamber) that prevent proper imaging of the iris & 36 & 207\\
\hline
\end{tabular}
\end{table*}

\subsection{Iris recognition tools}
This subsection provides a brief overview of iris recognition methods employed for the purpose of the experiments. We took effort to proceed with this study using four well-known, both commercial and academic iris recognition solutions.
 
\runinhead{\textbf{OSIRIS}} (\emph{Open Source for IRIS}) \cite{OSIRIS} is an open-source method developed in the framework of the BioSecure project, which follows Daugman's original concept of iris recognition based on the quantization of Gabor filtering outcomes to create a binary iris code. Iris codes are then compared against each other using the exclusive or (XOR) operation to yield a dissimilarity metric in a form of fractional Hamming distance normalized to a range of [0; 1], where values close to 0 are expected when samples coming from the same eye are compared, and values close to 0.5 would typically be expected for different-eye comparisons, as in the outcome of a series of independent coin tosses. However, as a result of the iris code shift to find the best match, intended as a countermeasure against eyeball rotation, impostor comparison scores distribution is usually centered around 0.4--0.45 values.

\runinhead{\textbf{VeriEye}} is a commercially available product, offered by Neurotechnology \cite{VeriEye}. Its encoding methodology is not disclosed by the manufacturer, apart from the claim that the method employs non-circular approximations of the iris boundaries using active shape modeling. This matcher gives comparison scores in the form of a similarity metric, \ie the greater the score, the better the match. Scores near zero (typically lower than 40) are considered different-eye comparisons, while scores above 40 are by default classified as same-eye comparisons. The highest recorded in our experiments score (when two identical images are compared) is 1557. 

\runinhead{\textbf{IriCore}} is another commercially available matcher, offered by IriTech Inc. \cite{IriCore}. Similarly to the VeriEye method, its manufacturer does not divulge the implementation details. Comparison scores when same-eye images are matched against each other are expected to fall in the range of 0 to 1.1, while different-eye comparisons should yield scores between 1.1 and 2.0.   

\runinhead{\textbf{MIRLIN}} is a method also offered on the market in the form of an SDK \cite{MIRLIN} by Fotonation Ltd. (formerly Smart Sensors Ltd.). The underlying methodology of this product is said to incorporate discrete cosine transform (DCT) applied to overlapping iris image patches in order to calculate binary code out of iris features \cite{Monro2007}. Binary templates are later on compared with each other to return fractional Hamming distance as a dissimilarity metric.

\subsection{Examination of cataract influence}
\begin{figure}[htb]
\begin{center}
	\includegraphics[width=0.497\linewidth]{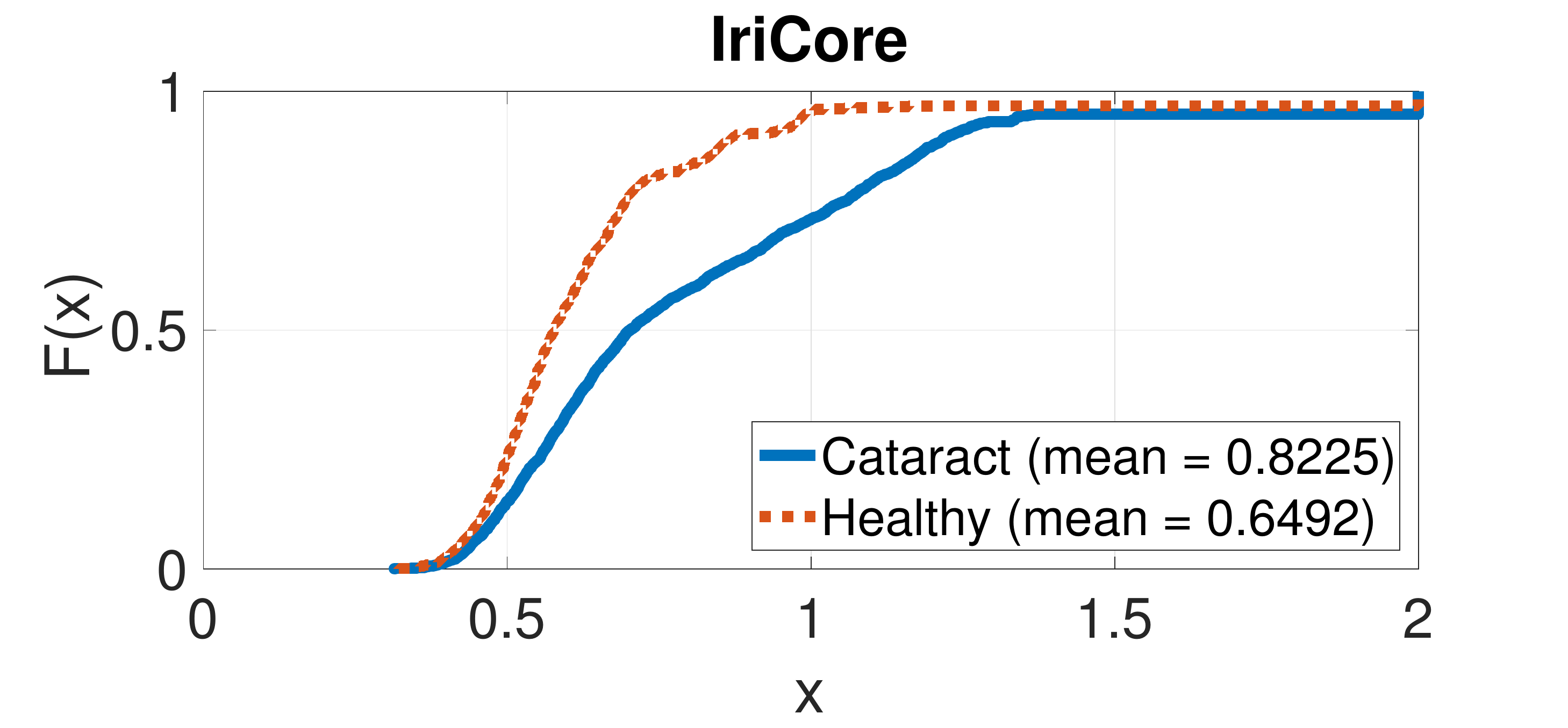}
	\hfill
	\includegraphics[width=0.497\linewidth]{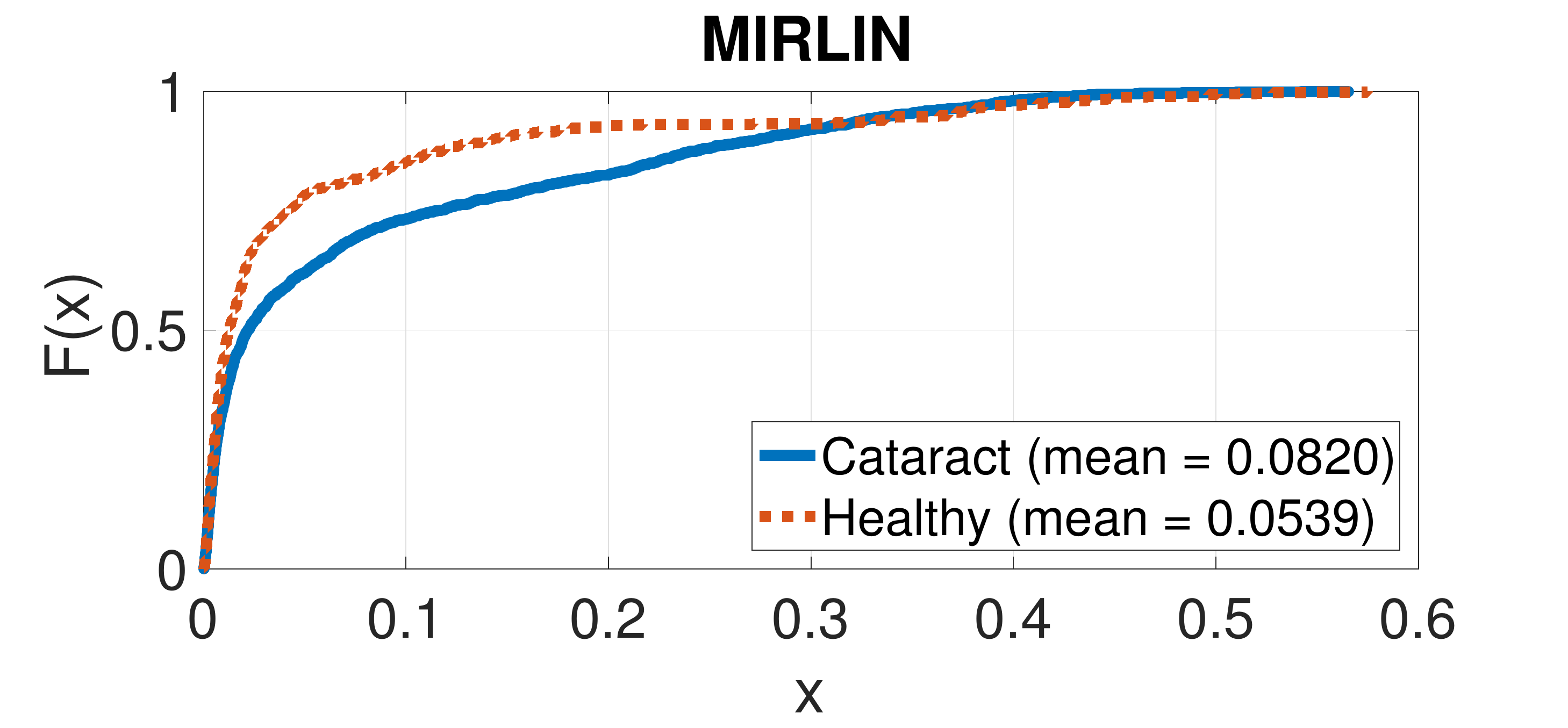}
	\vfill\vskip2mm
	 \includegraphics[width=0.497\linewidth]{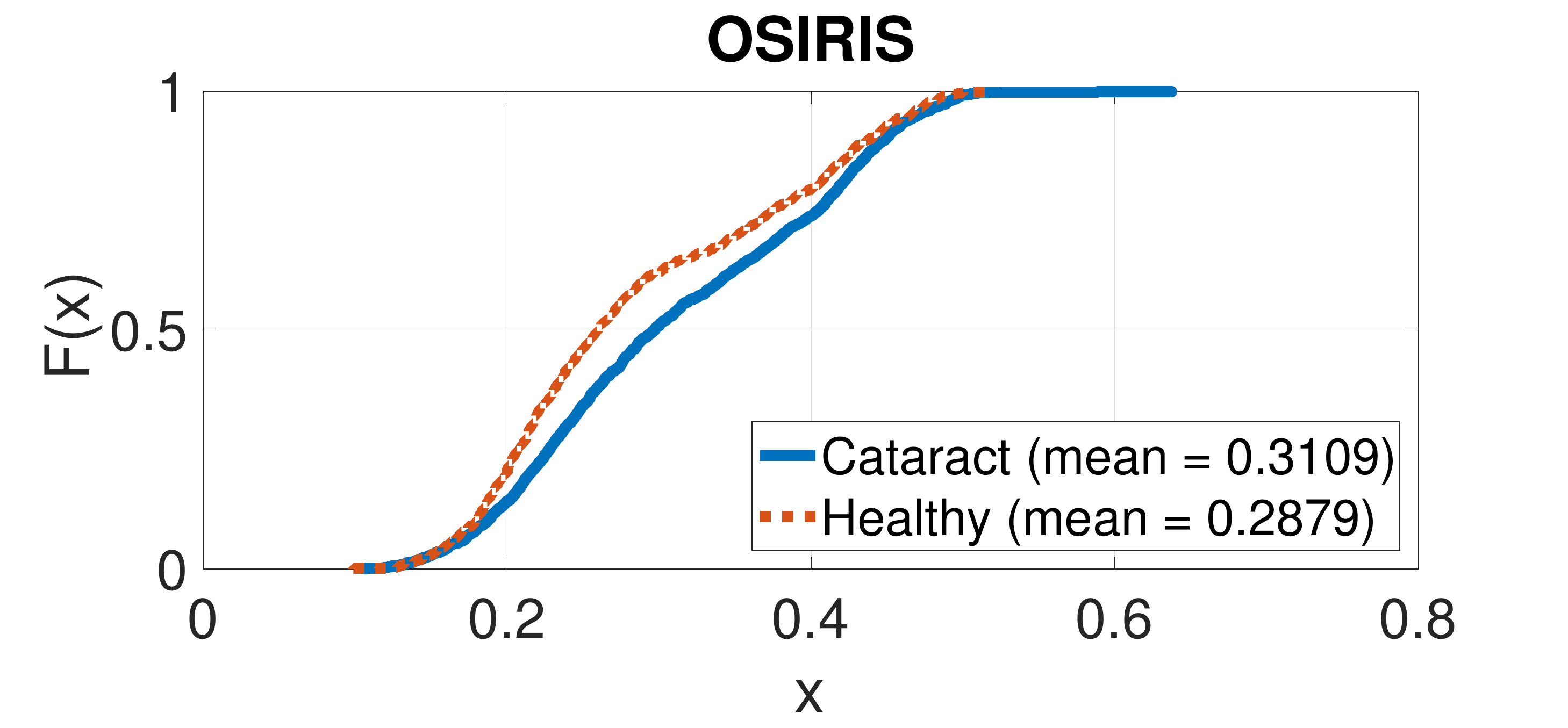}
  \hfill
  \includegraphics[width=0.497\linewidth]{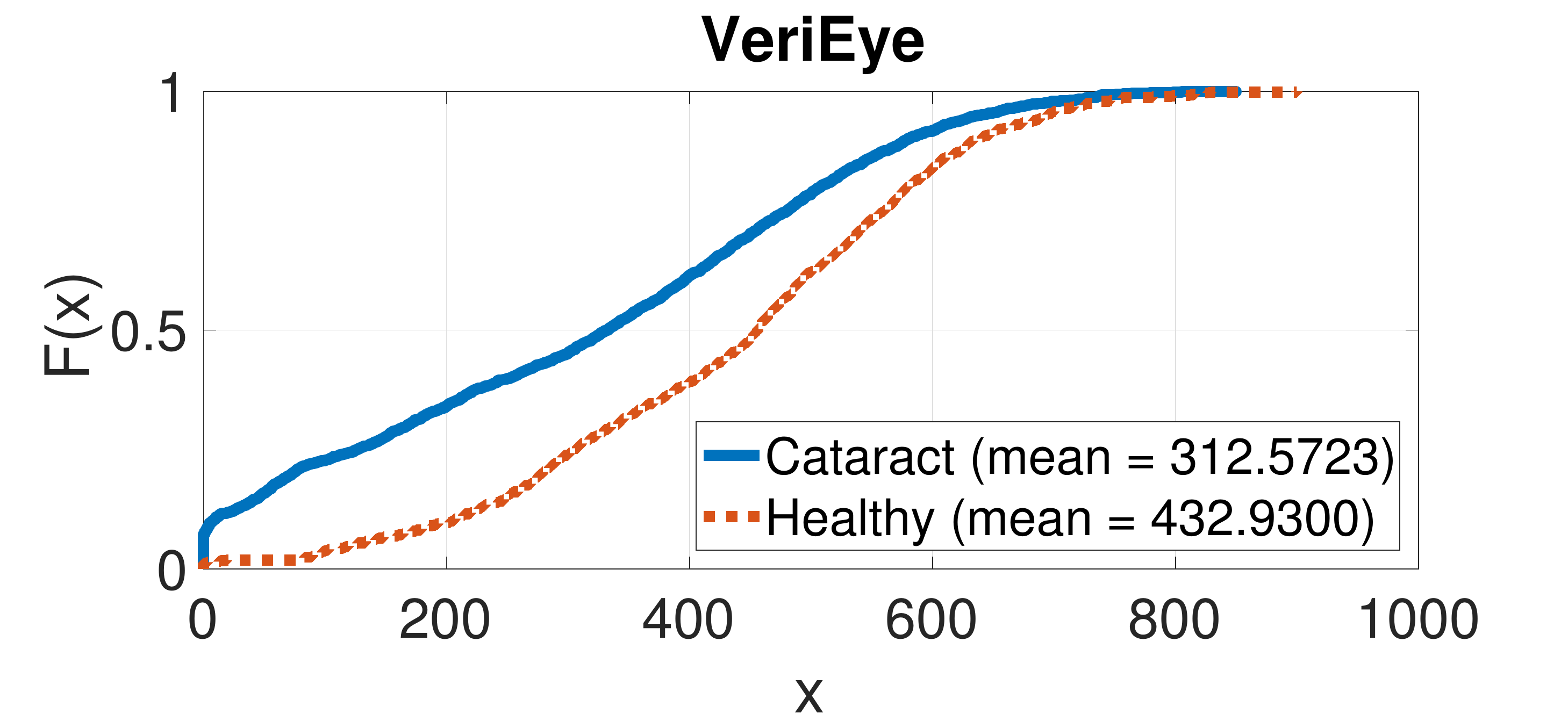}
\end{center}
\vskip-3mm
\caption{Cumulative distribution functions for \textbf{genuine comparisons} obtained for all four iris recognition methods denoting the performance of these systems when cataract eyes are enrolled compared to a control group of healthy eyes. Mean values are provided in brackets.}
\label{fig::cataract-F-gens}
\end{figure}

\begin{figure}[htb]
\begin{center}
	\includegraphics[width=0.497\linewidth]{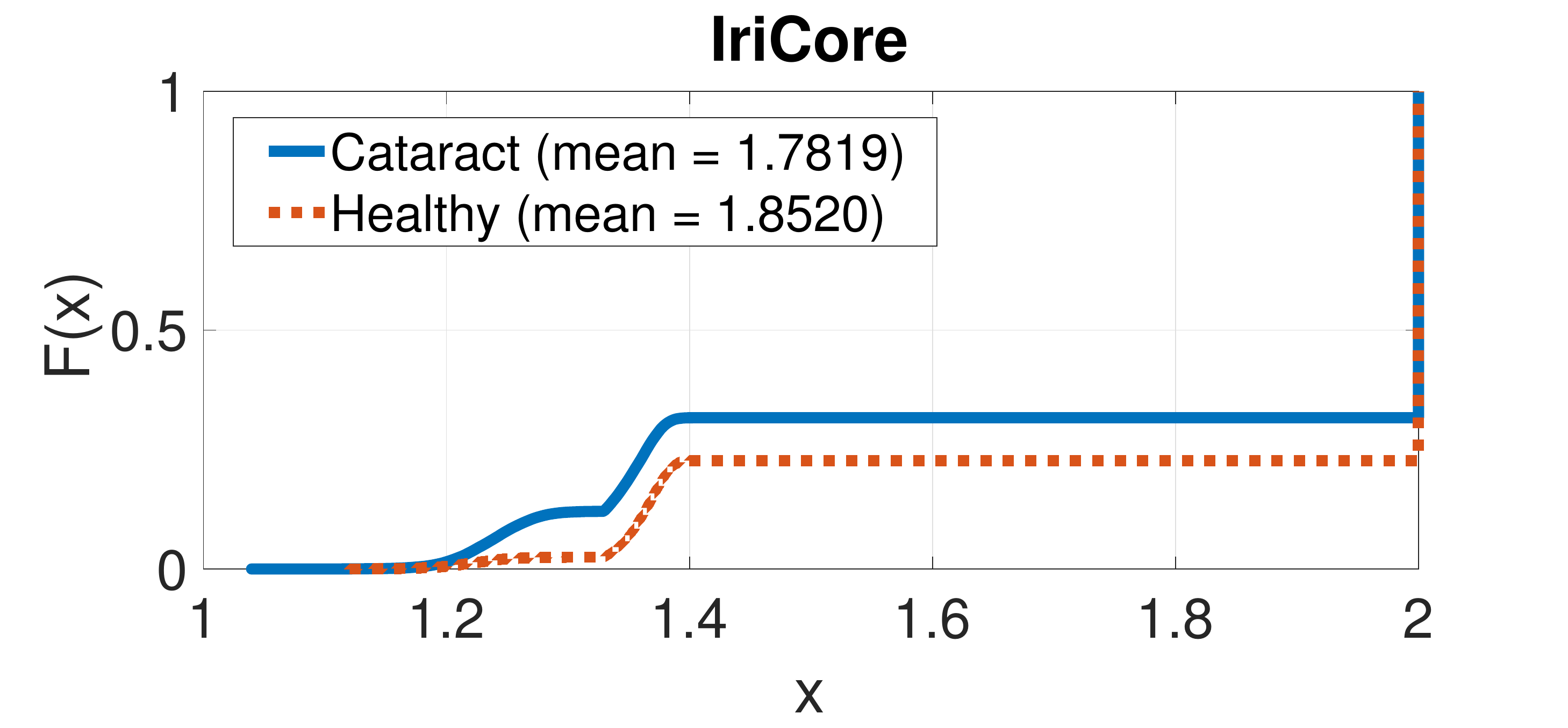}
	\hfill
	\includegraphics[width=0.497\linewidth]{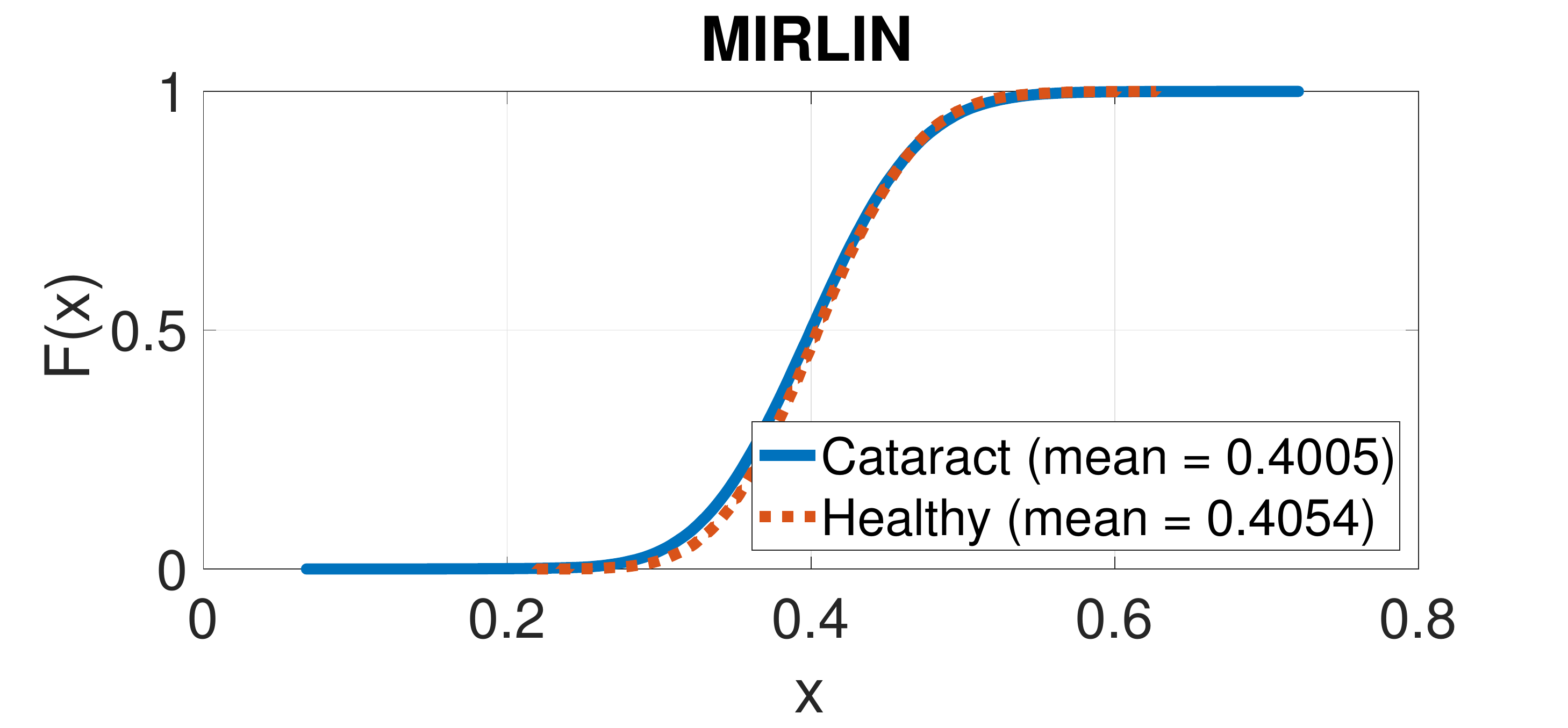}
	\vfill\vskip2mm
	 \includegraphics[width=0.497\linewidth]{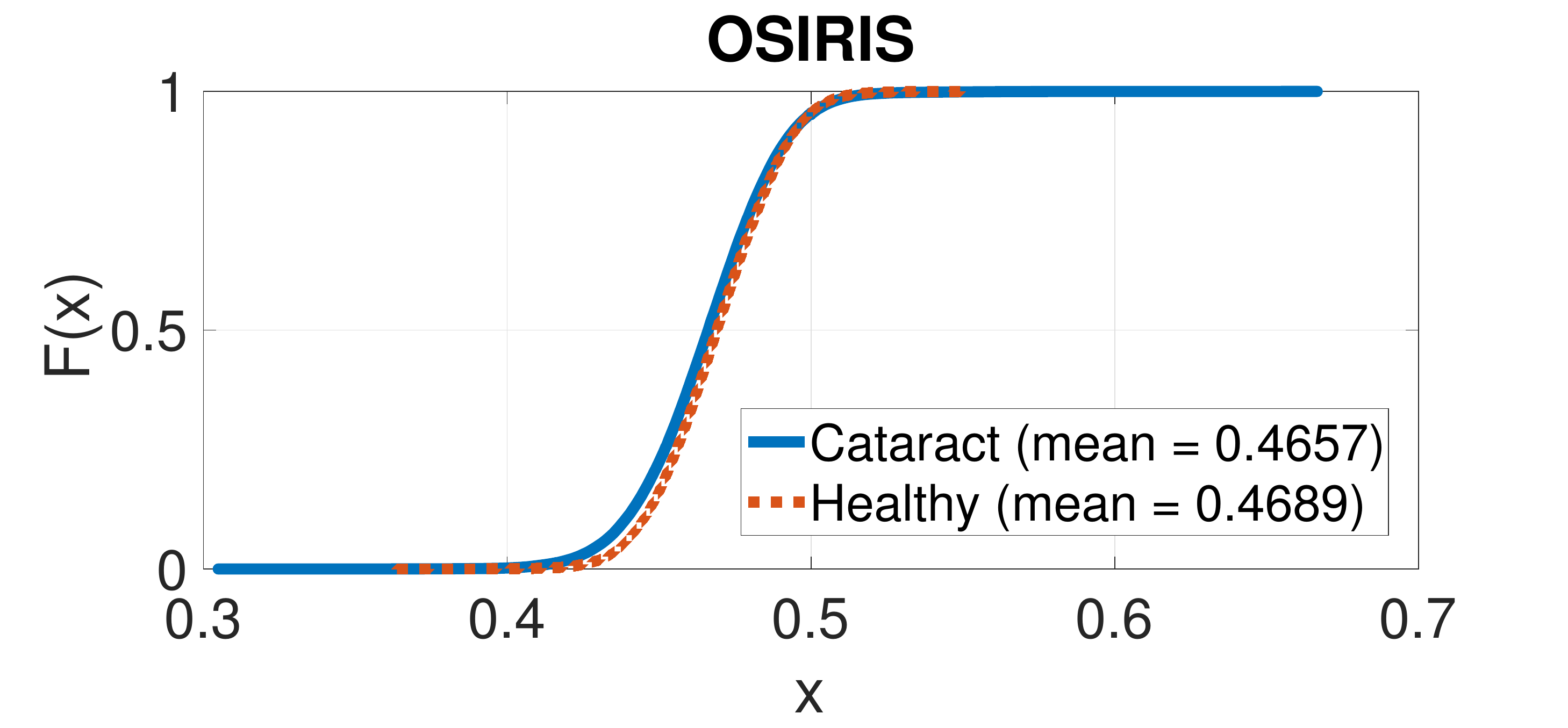}
  \hfill
  \includegraphics[width=0.497\linewidth]{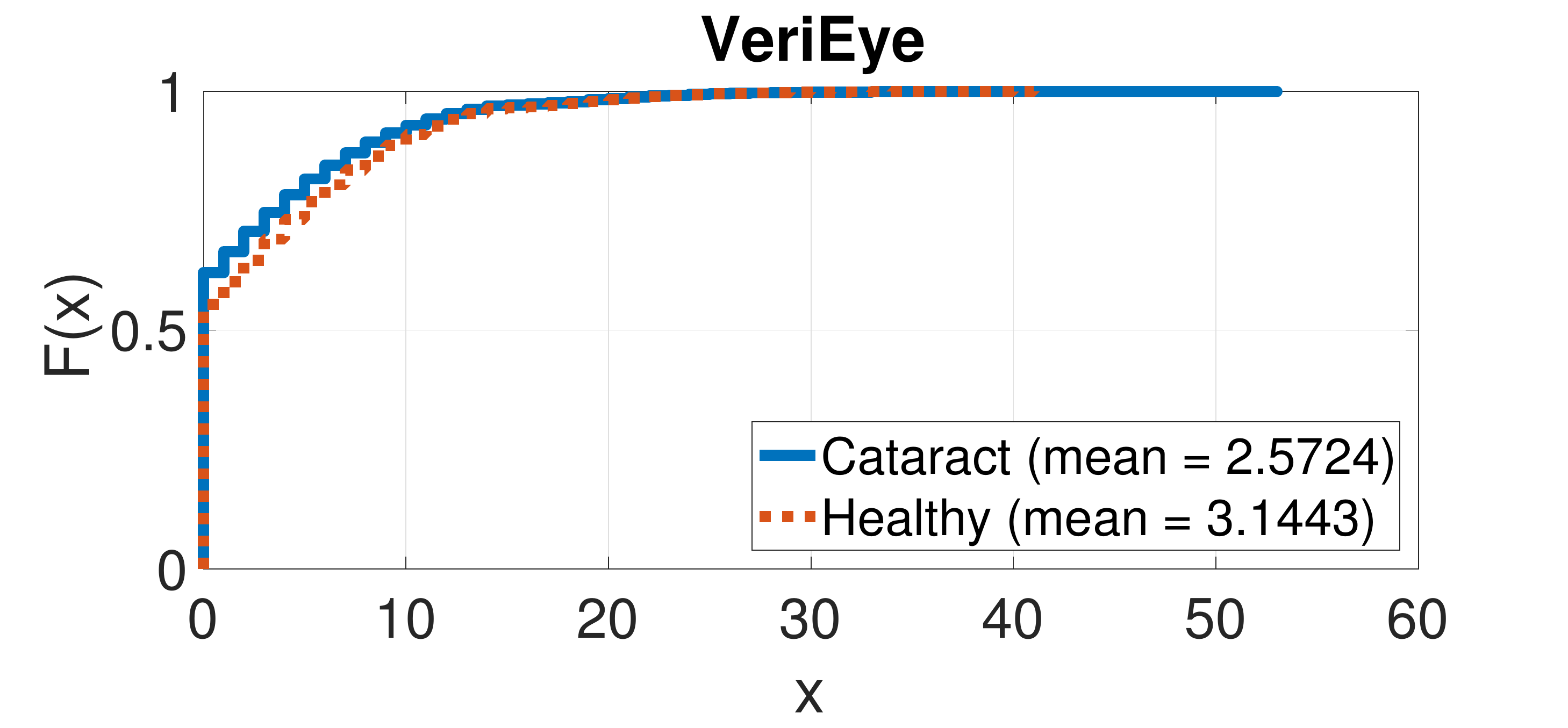}
\end{center}
\vskip-3mm
\caption{Same as in Fig. \ref{fig::cataract-F-gens}, except that graphs for \textbf{impostor comparisons} are presented, together with respective mean values in brackets.}
\label{fig::cataract-F-imps}
\end{figure}

For both the \emph{Cataract} and \emph{Healthy} subsets, all possible genuine and impostor comparisons were performed to come up with comparison score distributions in the form of cumulative distribution functions (CDFs), which are to reveal whether cataract-afflicted eyes are expected to perform worse than their healthy counterparts, Figs. \ref{fig::cataract-F-gens} and \ref{fig::cataract-F-imps}. Notably, for each of the four iris matchers we can observe a visible shift of the genuine score distribution obtained from the \emph{Cataract} subset towards worse scores (\ie lower for VeriEye, and higher for the remaining three methods). Regarding impostor scores, the differences in score distributions are smaller and uneven across matchers (highest for IriCore, negligible for MIRLIN and OSIRIS).  

To examine whether the observed differences can be considered statistically significant, a two-sample Kolmogorov-Smirnov test is applied with the significance level $\alpha=0.05$ (further referred to as K-S test). One-sided variant of the test is used when comparing genuine score distributions, and a two-sided variant for the impostor score distributions. The K-S test makes no assumptions on the distributions (apart from their continuity) and the test statistics simply quantifies the distance between two empirical cumulative distribution functions $F(x_1)$ and $F(x_2)$ of the random variables $x_1, x_2$ being compared. To alleviate the issue of statistical dependencies between comparison scores that are introduced when performing all possible comparisons between samples, we resample with replacement each set of comparison scores 1,000 times for genuine scores and 10,000 times for impostor scores, providing sets of statistically independent scores. These resampled sets of scores are later used for performing the K-S tests, whose results are presented in Table \ref{table:cataract-KS-tests}. 

\begin{table}[h!]
\renewcommand{\arraystretch}{1.1}
\caption{Kolmogorov-Smirnov statistical testing results for the cataract experiment. The null hypotheses $H_0$ in all tests state that the scores originating from two compared partitions are drawn from the same distribution. Alternative hypotheses $H_1$ for genuine scores state that scores obtained from \emph{Cataract} set are {\bf worse} than those obtained from \emph{Healthy} partition, while for impostor scores $H_1$ state that scores obtained from \emph{Cataract} set are {\bf different} than those obtained from \emph{Healthy} partition One-sided test is used for genuine comparisons and two-sided test for impostor comparisons.}
\label{table:cataract-KS-tests}
\centering\footnotesize
\begin{tabular}[t]{c|c|c|}
\cline{2-3}
& \textbf{Genuine comparisons} & \textbf{Impostor comparisons} \\ 
& \emph{Cataract} ($g_c$) vs. \emph{Healthy} ($g_h$) & \emph{Cataract} ($i_c$) vs. \emph{Healthy} ($i_h$) \\
& \textbf{$H_1: F(g_c) < F(g_h)$} & $H_1: F(i_c) \nsim F(i_h)$ \\\hline\hline
{\bf OSIRIS} & $<0.0001$ & $<0.0001$  \\\hline
{\bf MIRLIN} &  $<0.0001$ & $<0.0001$ \\\hline
{\bf IriCore} & $<0.0001$ & $<0.0001$ \\\hline
& \emph{Cataract} ($g_c$) vs. \emph{Healthy} ($g_h$) & \emph{Cataract} ($i_c$) vs. \emph{Healthy} ($i_h$) \\
& $H_1: F(g_c) > F(g_h)$ & $H_1: F(i_c) \nsim F(i_h)$ \\\hline
{\bf VeriEye} & $<0.0001$ & $<0.0001$ \\\hline
\end{tabular}
\end{table}

\begin{figure}[htb]
\begin{center}
	\includegraphics[width=0.497\linewidth]{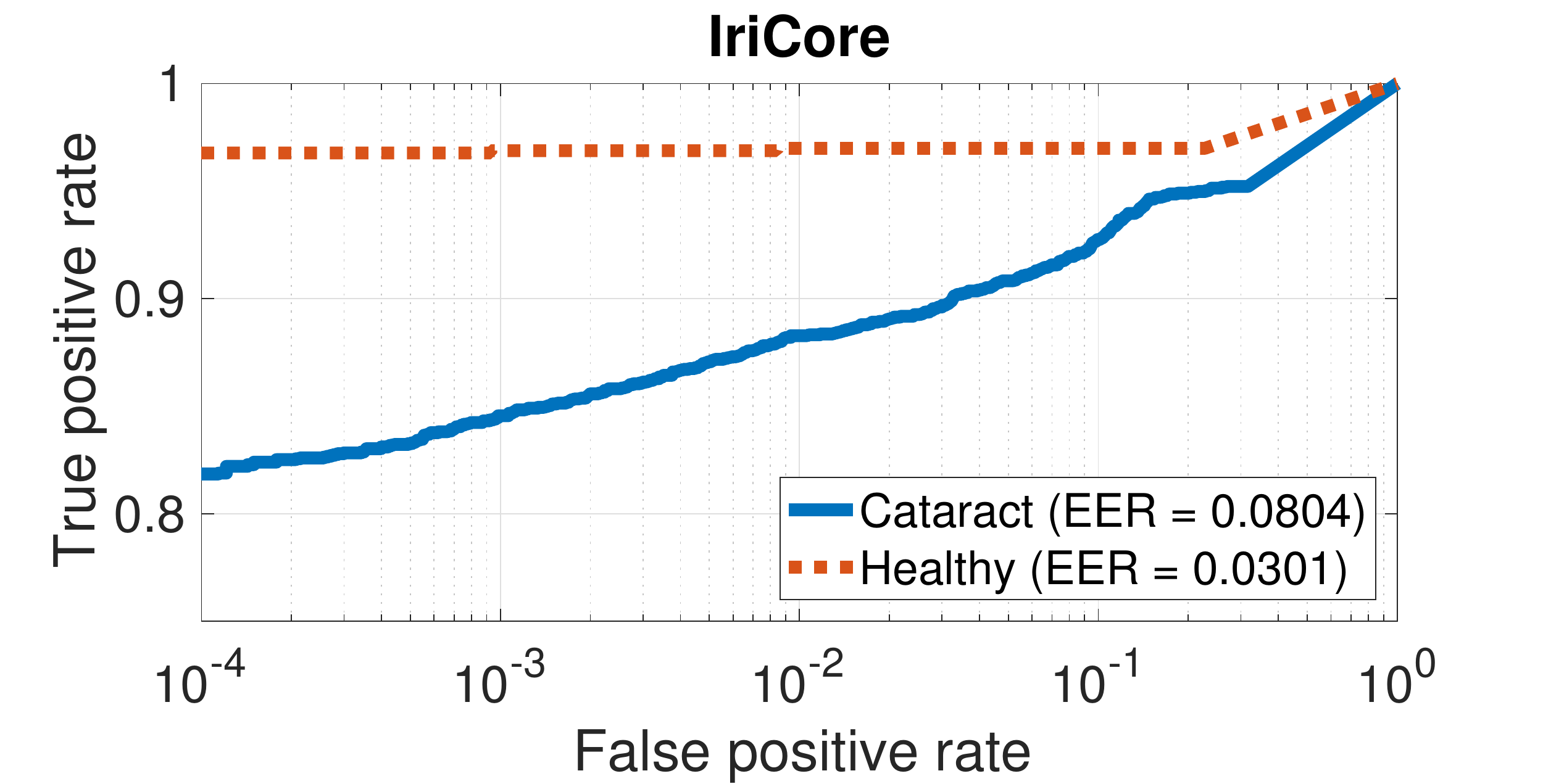}
	\hfill
	\includegraphics[width=0.497\linewidth]{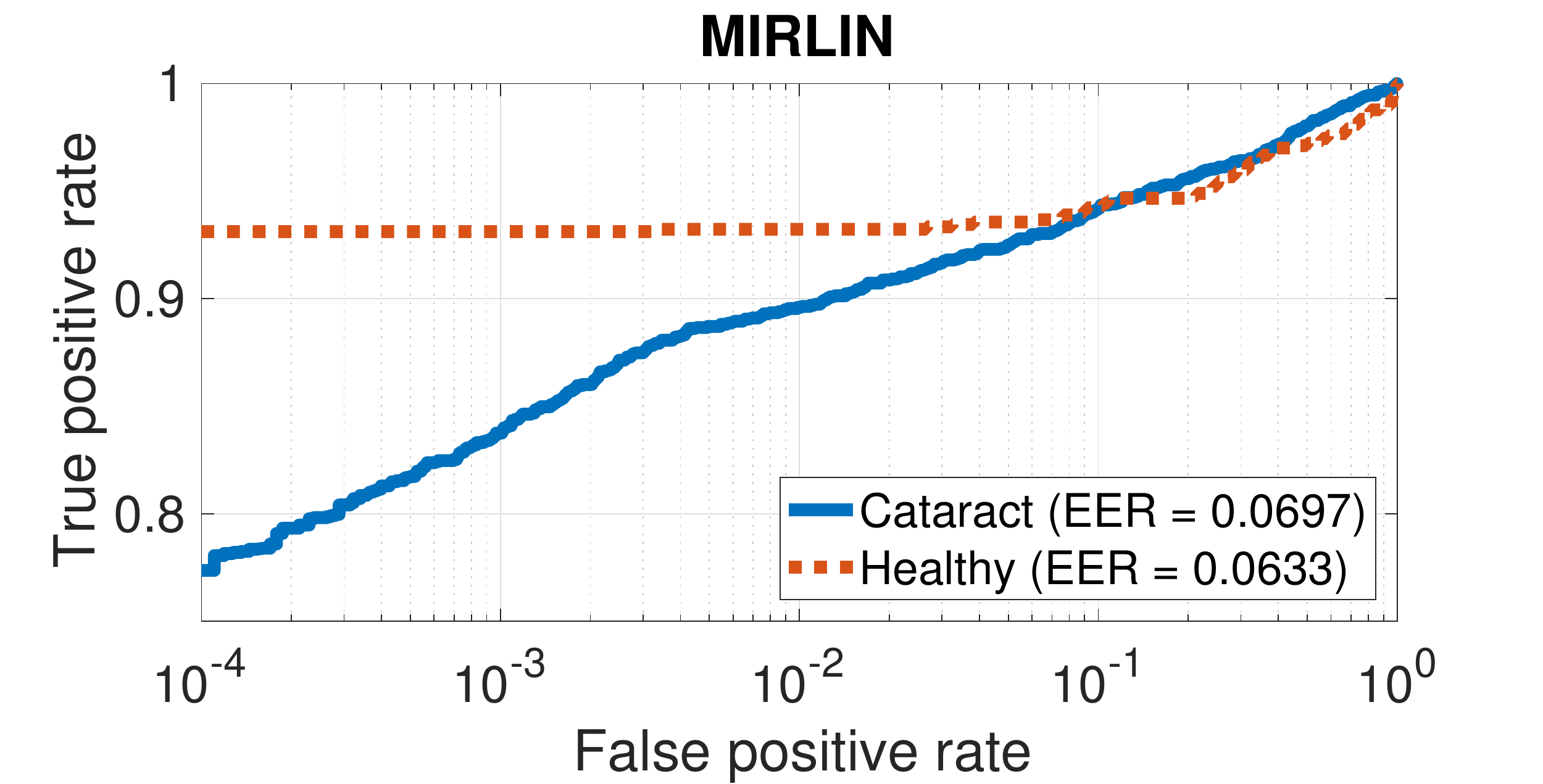}
	\vfill\vskip2mm
	 \includegraphics[width=0.497\linewidth]{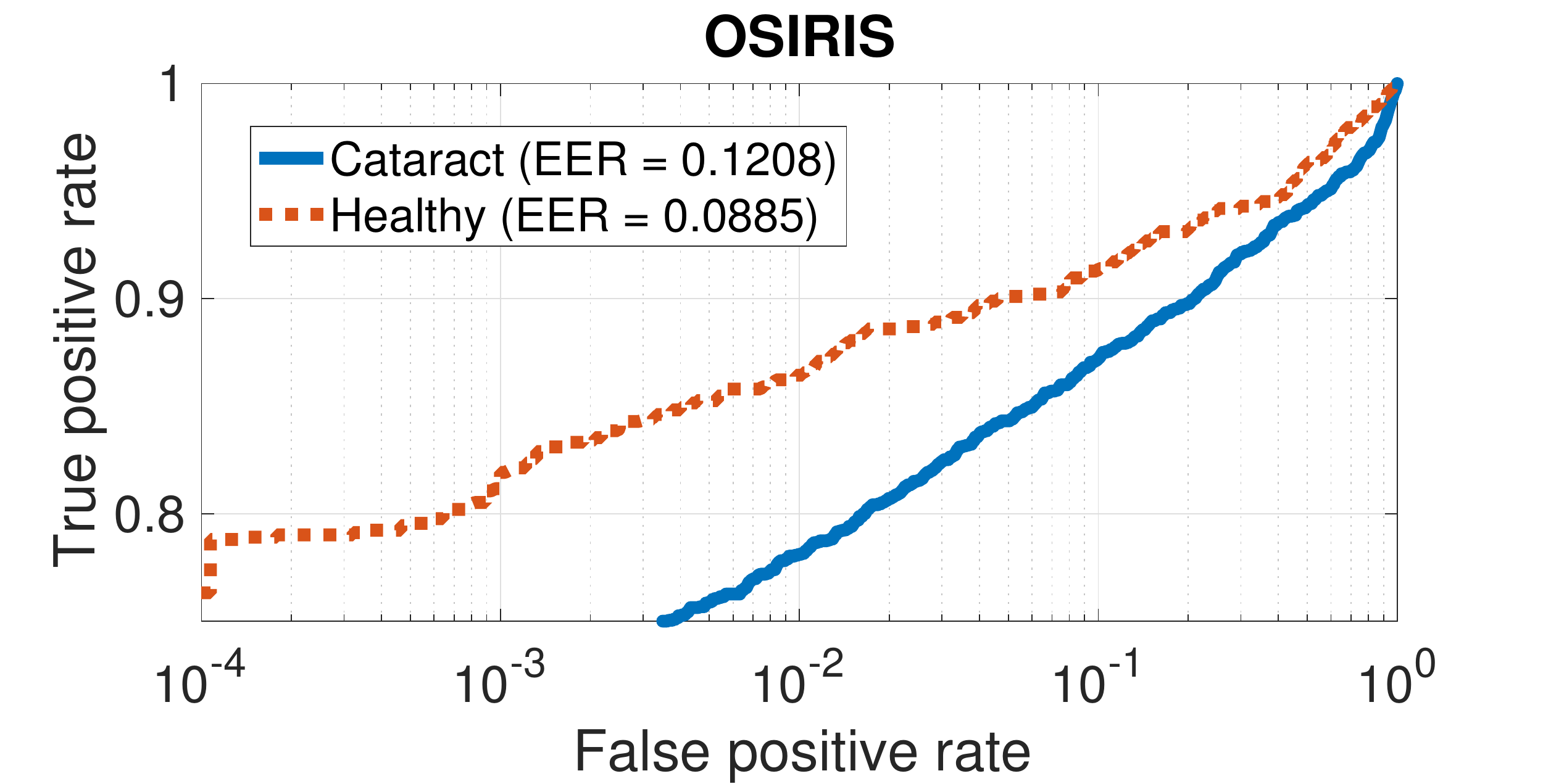}
  \hfill
  \includegraphics[width=0.497\linewidth]{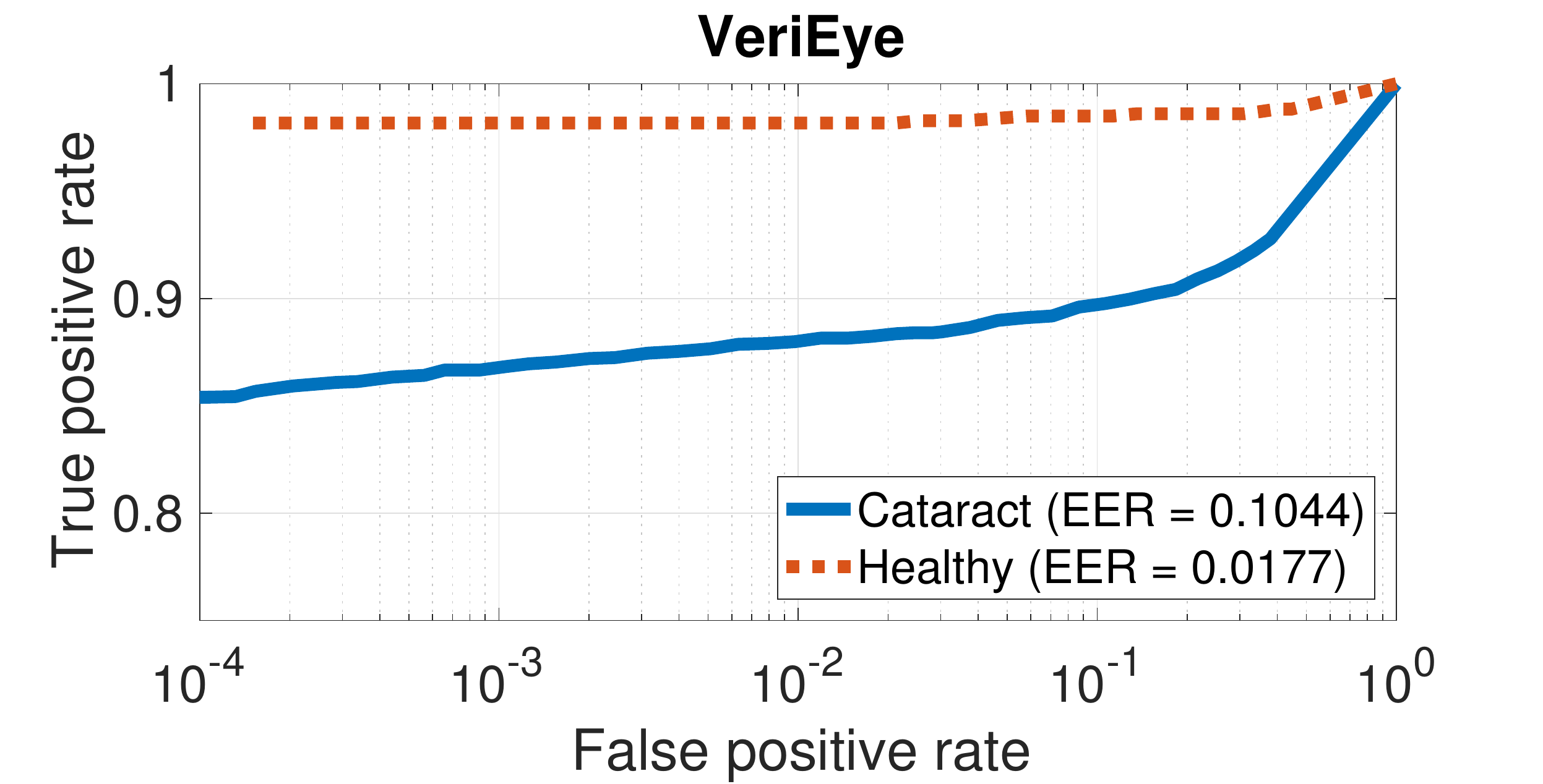}
\end{center}
\vskip-3mm
\caption{Receiver Operating Characteristic (ROC) curves obtained for all four iris recognition methods denoting the performance of these systems when cataract eyes are enrolled compared to a control group of healthy eyes. Equal Error Rate values are provided in brackets.}
\label{fig::cataract-ROCs}
\end{figure}

In addition to cumulative distribution function graphs, we also present Receiver Operating Characteristic (ROC) curves which demonstrate the accuracy of these iris recognition systems when they are presented with eyes afflicted by cataract, compared to a scenario, in which healthy eyes are used. The ROC curve presents a relation of true positive ratio to false positive ratio obtained by a given decision system and is therefore helpful for assessing its expected behavior. ROCs for respective iris matchers are shown in Fig. \ref{fig::cataract-ROCs}. Equal Error Rate values are also provided as another metric for quantifying a system's performance. For all four employed iris recognition methods, the \emph{Cataract} subset gives worse ROC-wise performance that the \emph{Healthy} subset except for the MIRLIN matcher, which gives similar EER values for both partitions. 

\pagebreak

\subsection{Examination of disease influence by types of eye damage}
\begin{figure}[htb]
\begin{center}
	\includegraphics[width=0.497\linewidth]{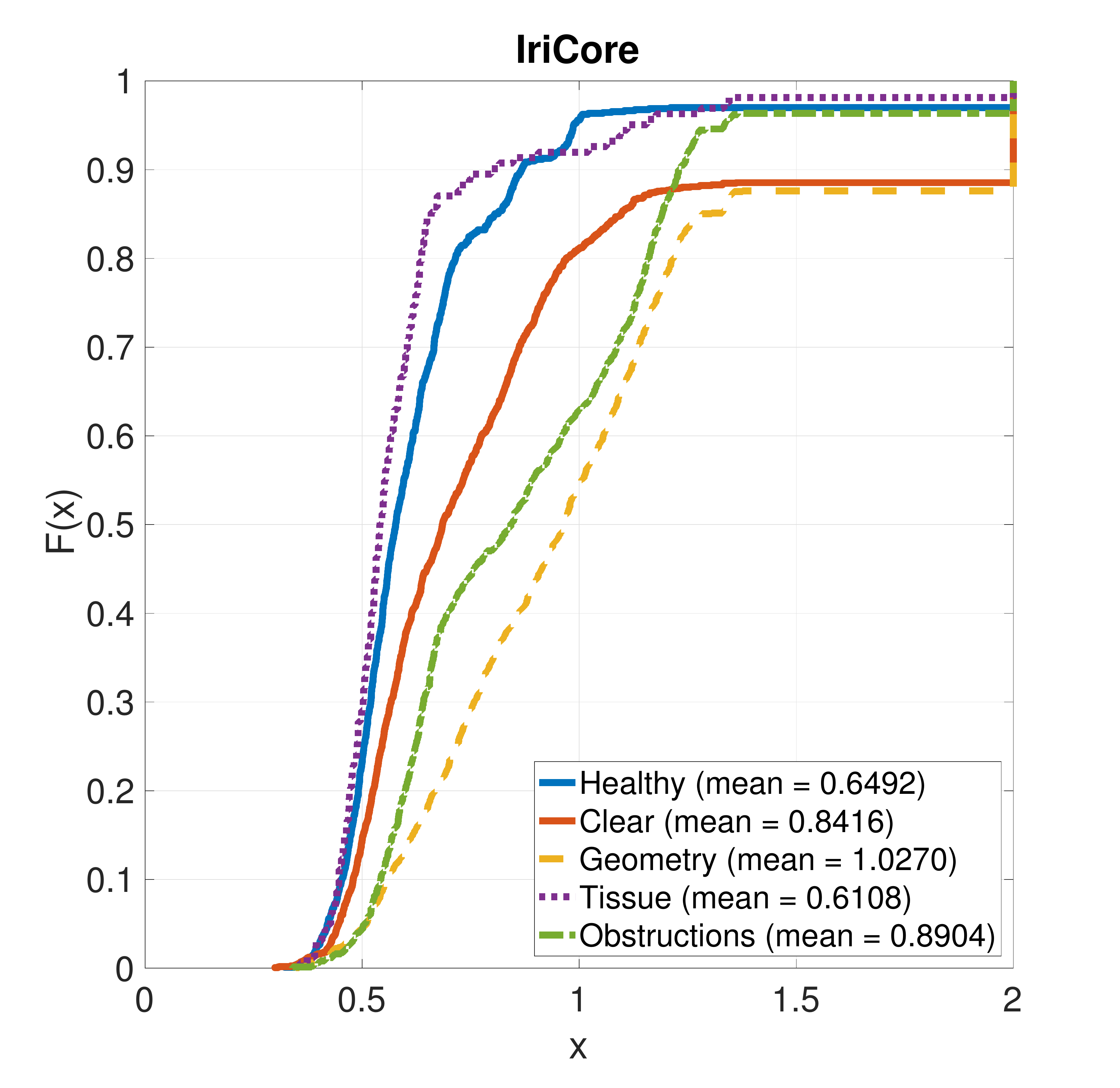}
	\hfill
	\includegraphics[width=0.497\linewidth]{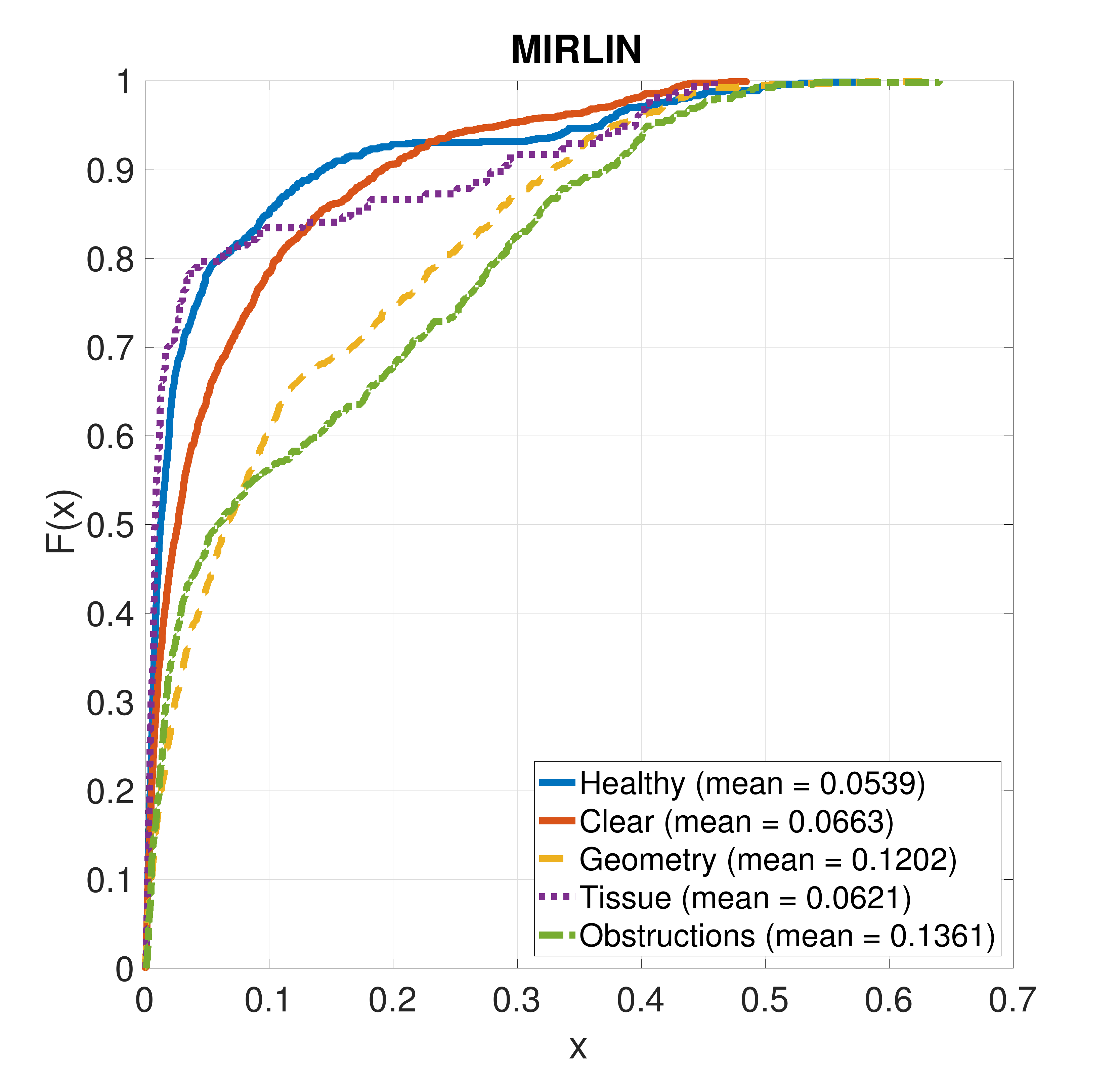}
	\vfill
	 \includegraphics[width=0.497\linewidth]{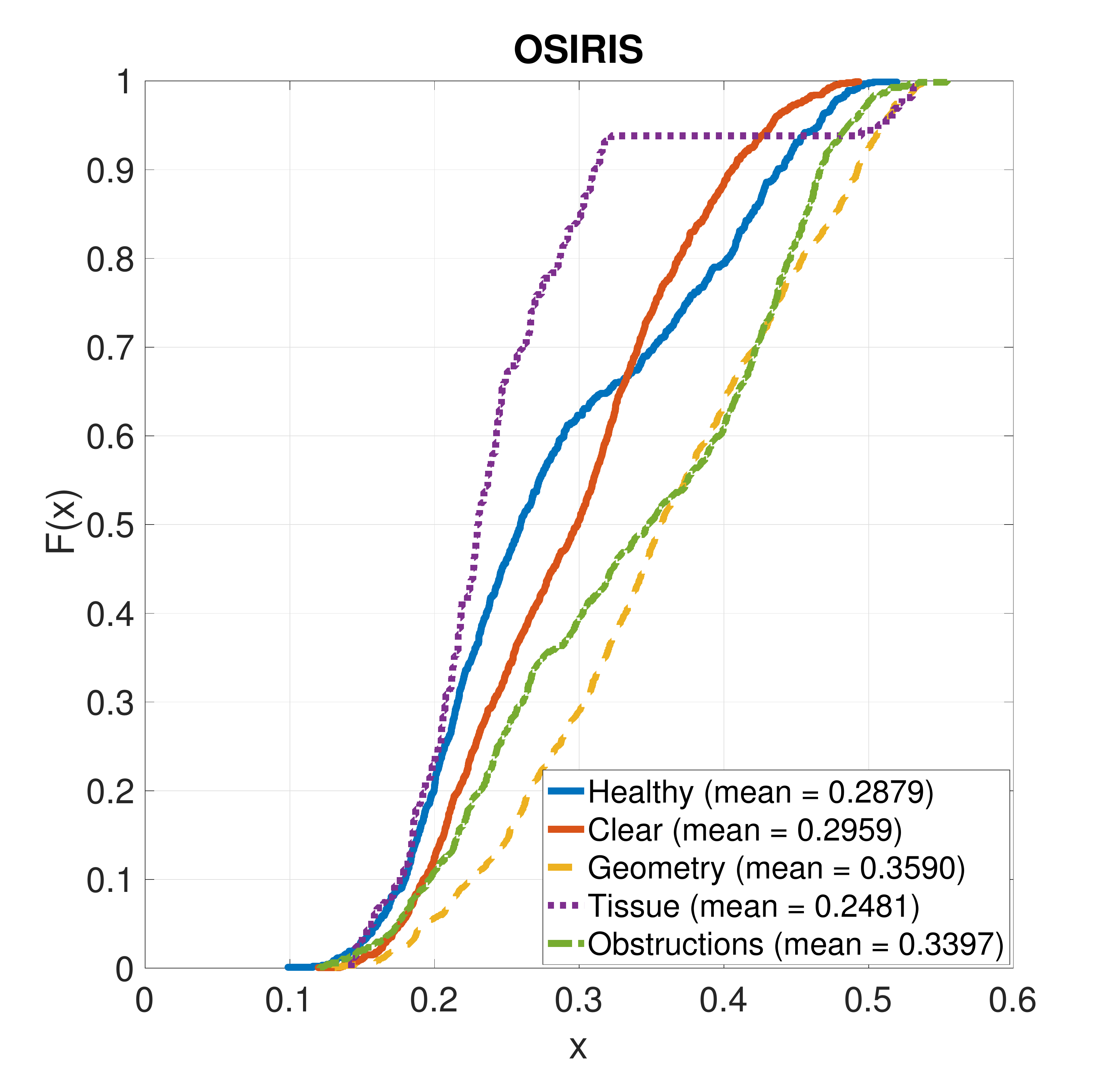}
  \hfill
  \includegraphics[width=0.497\linewidth]{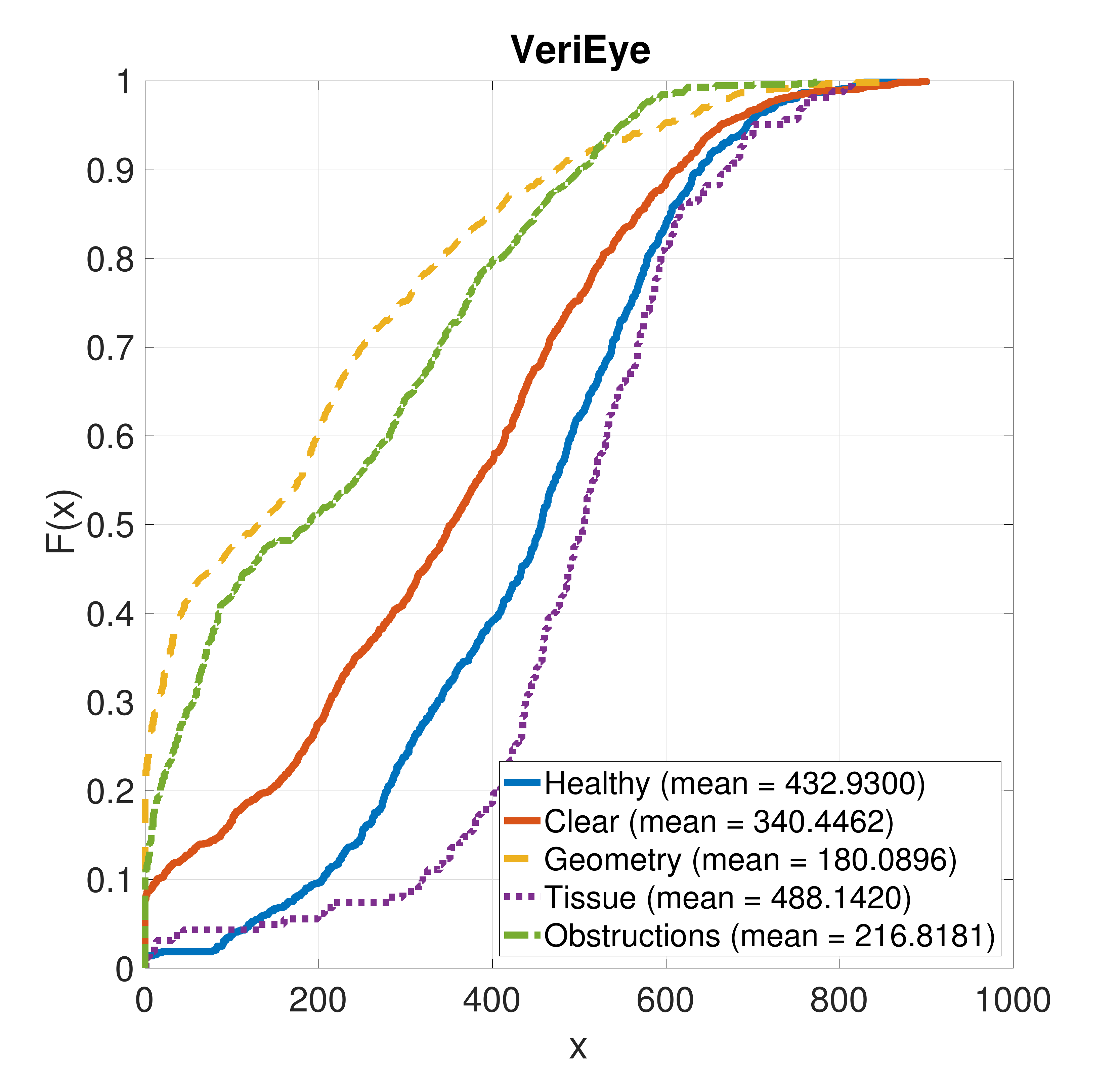}
\end{center}
\vskip-3mm
\caption{Cumulative distribution function plots for \textbf{genuine comparisons} obtained for all four iris recognition methods denoting the performance of these systems for five data subsets. Mean values are provided in brackets. Note that these graphs may differ from those in \cite{TrokielewiczBTAS2015}, as here we use samples that are not required to comply to the ISO/IEC 29794-6 iris image quality standard.}
\label{fig::subsets-F-gens}
\end{figure}

Accordingly with the testing procedure described for the \emph{Cataract} subset experiments, in this section we present CDF graphs complemented by the results of the K-S statistical testing, together with ROC curves for all four iris recognition methods employed. The CDFs for genuine comparison scores are presented in Fig. \ref{fig::subsets-F-gens}. Similarly across all methods, the \emph{Geometry} and \emph{Obstructions} subsets present the worst scores, with the CDF shifted to the right for VeriEye, and to the left for the remaining matchers, when compared to the \emph{Healthy} subset serving as a control group. Uneven behavior can be observed for the \emph{Clear} subset, which gives worse scores for most matchers, except for the OSIRIS matcher, in which its CDF intertwines with the CDF corresponding to the \emph{Healthy} subset. Surprisingly, the \emph{Tissue} subset displays behavior that is either similar to this of the \emph{Healthy} subset, or even slightly better. These fluctuations can be, however, attributed to the small number of samples in the \emph{Tissue} subset, which makes it easier for given samples to influence the performance of the whole subset. Table \ref{table:statTestsGenuine} presents the results of K-S testing procedure for genuine comparisons, which confirms that the differences between the \emph{Clear}, \emph{Geometry}, and \emph{Obstructions} subsets when compared against the \emph{Healthy} subset are statistically significant. As for the impostor-related CDFs, again rather large differences may be observed for the IriCore matcher, and smaller for the remaining three methods. The K-S statistical testing again confirms that there are statistically significant differences between each of the four subsets comprising diseased eyes and the \emph{Healthy} partition, Tab. \ref{table:statTestsImpostor}).

\begin{figure}[htb]
\begin{center}
	\includegraphics[width=0.497\linewidth]{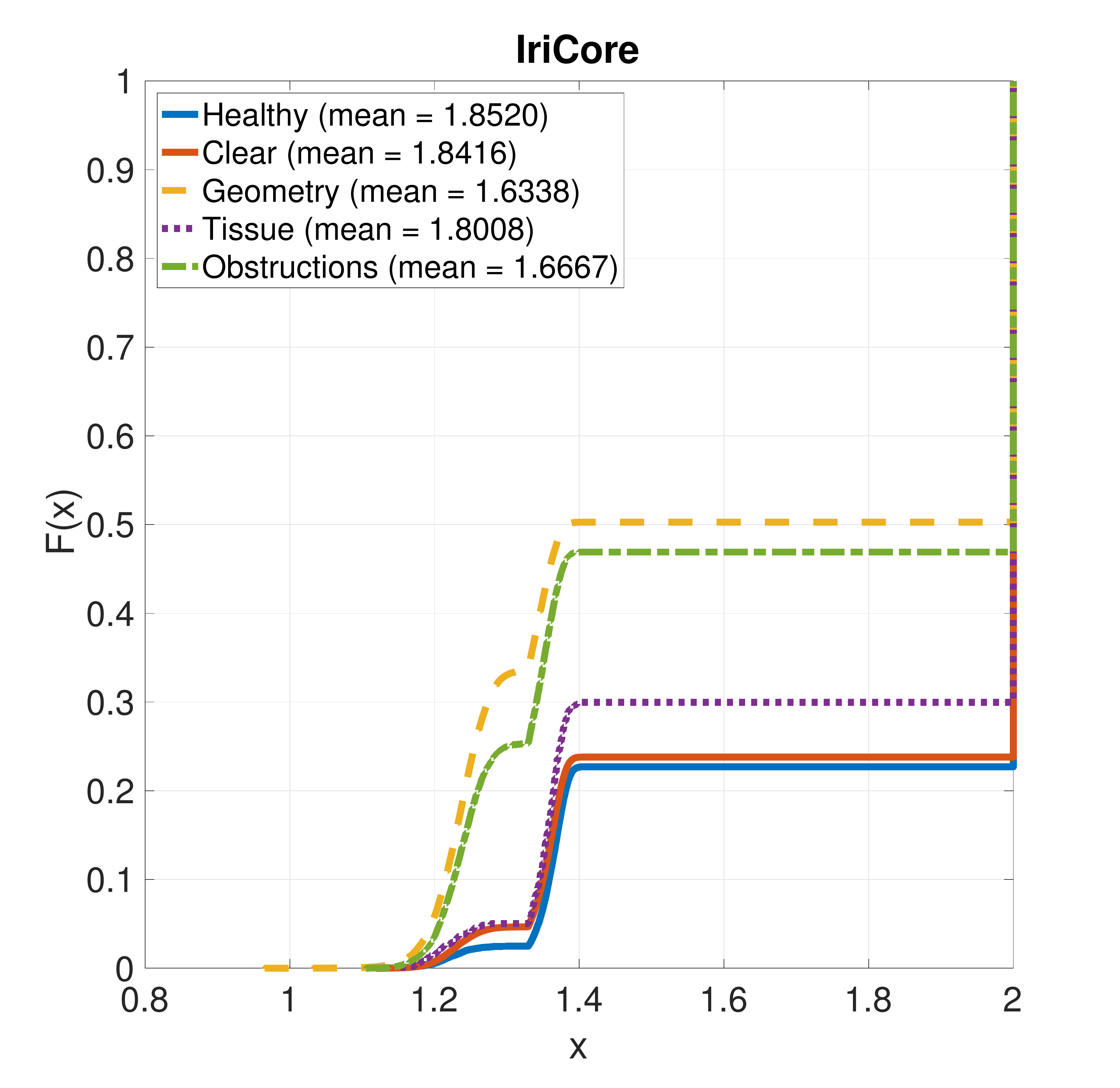}
	\hfill
	\includegraphics[width=0.497\linewidth]{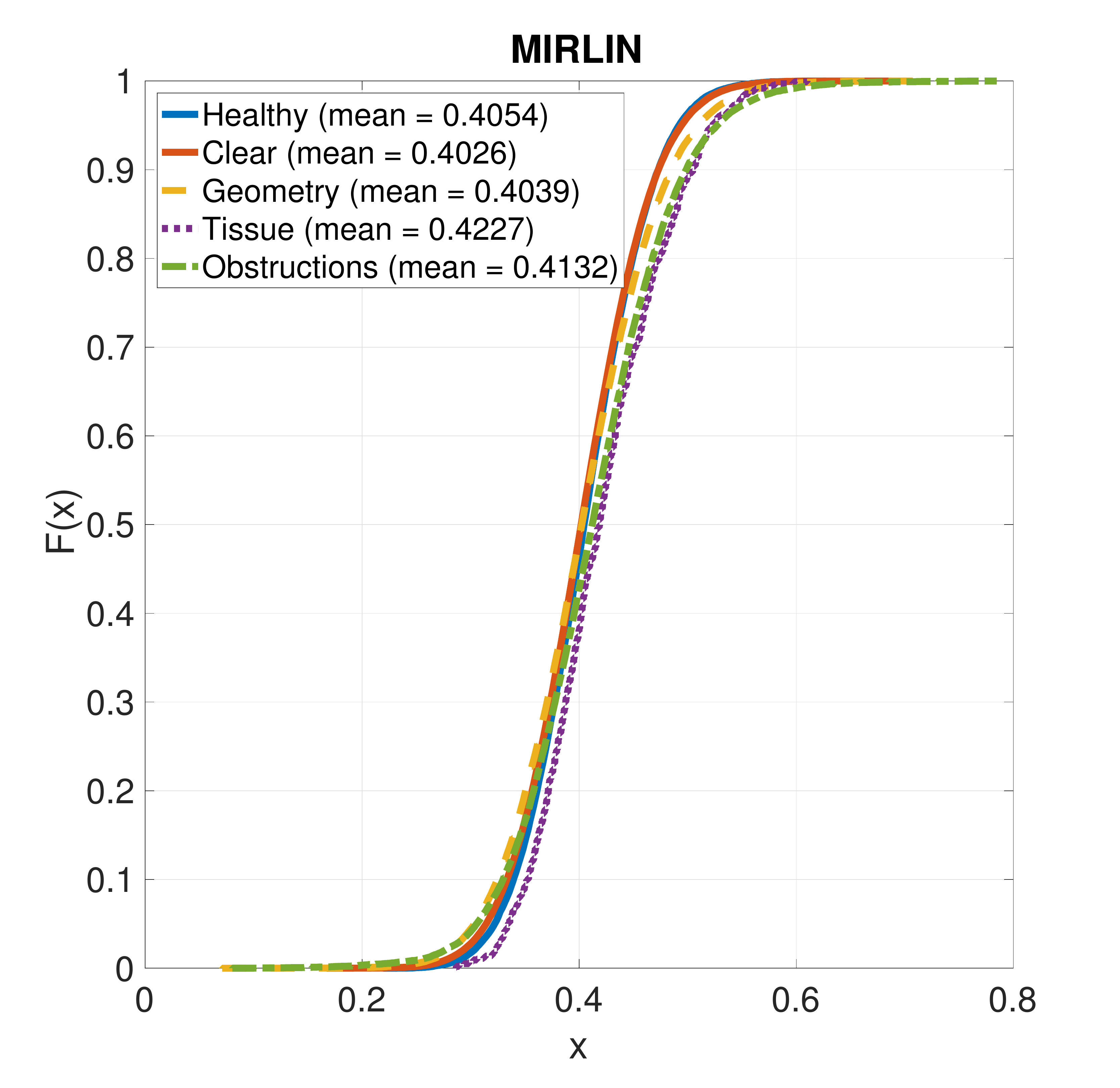}
	\vfill
	 \includegraphics[width=0.497\linewidth]{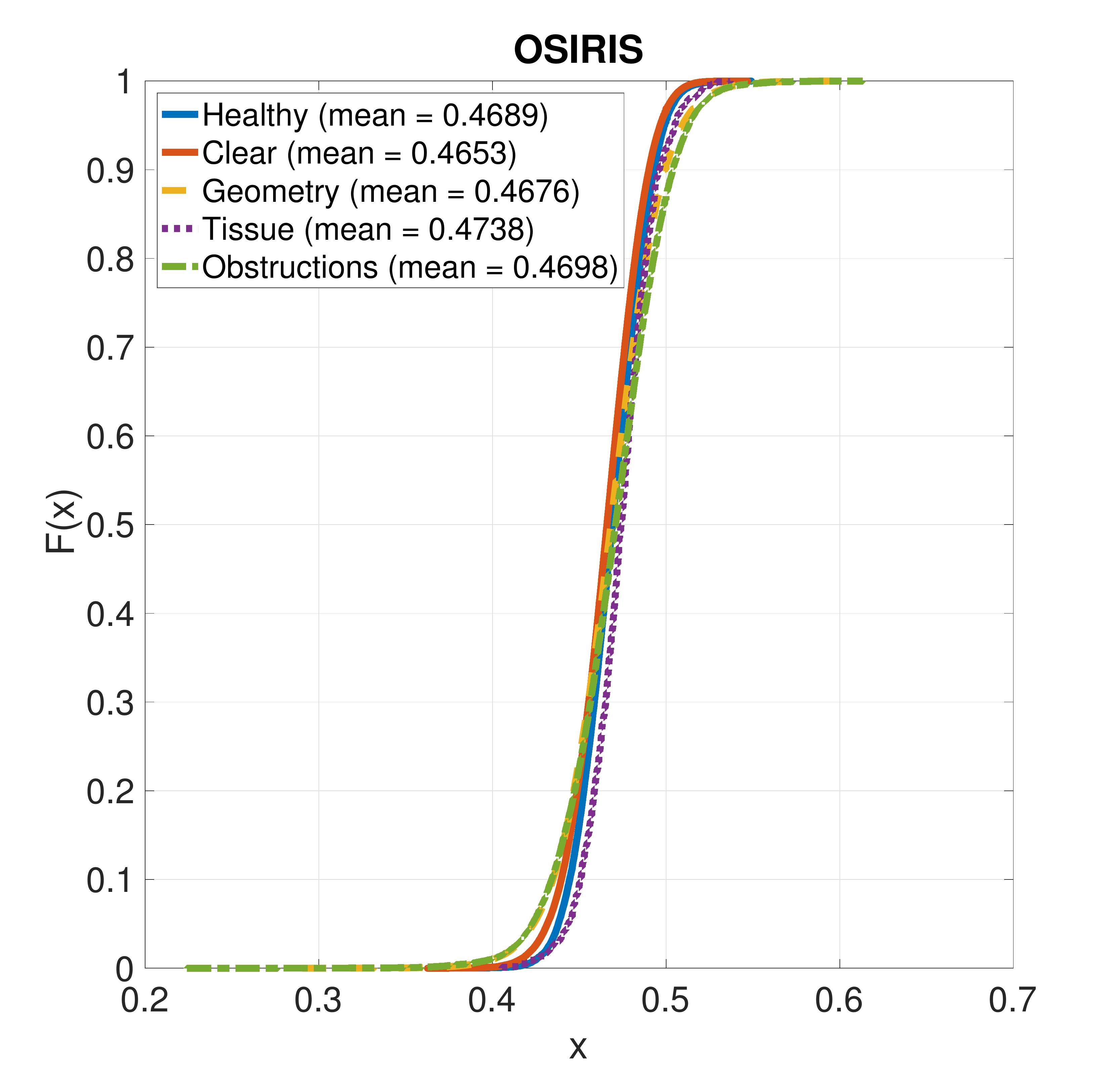}
  \hfill
  \includegraphics[width=0.497\linewidth]{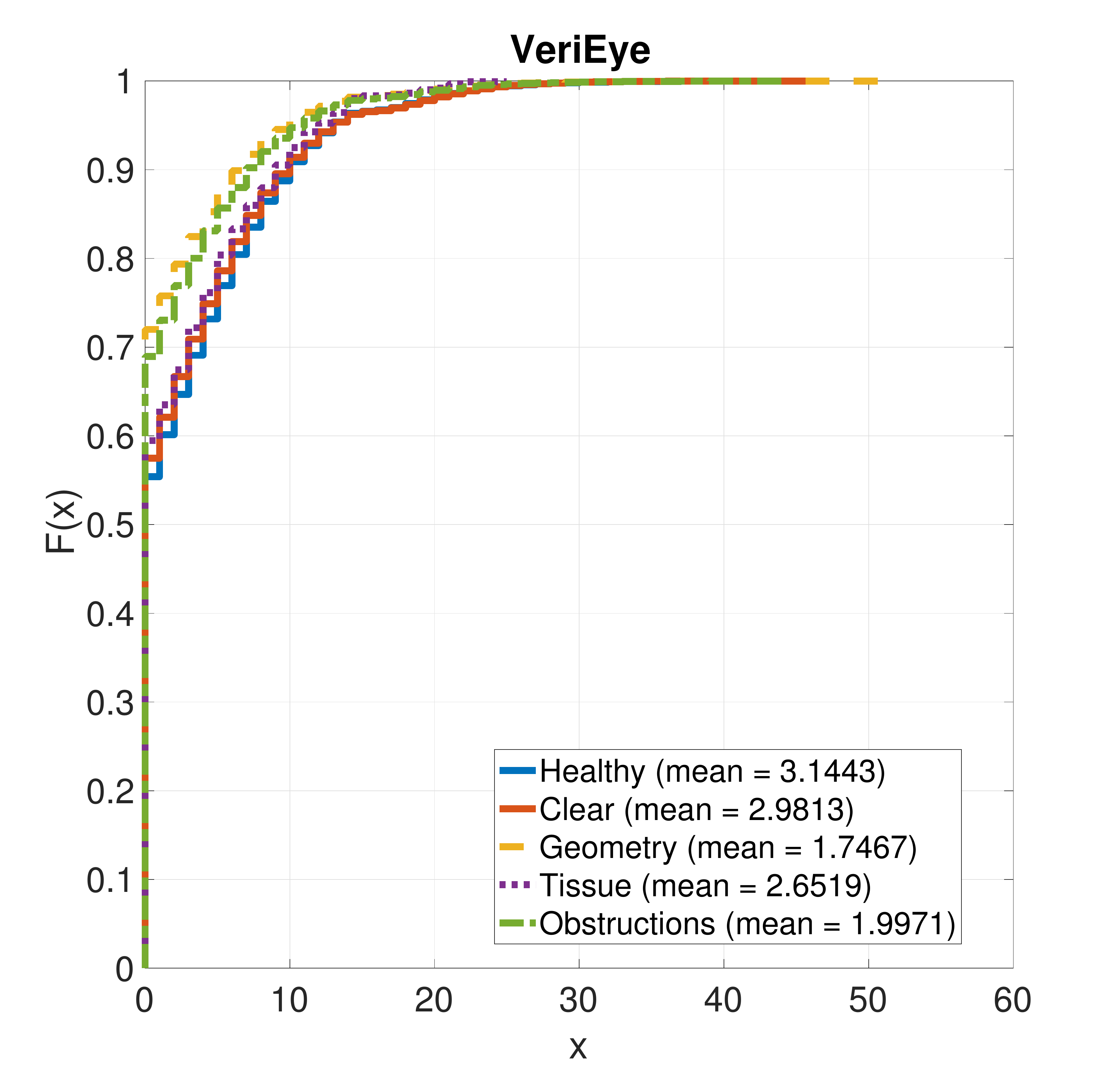}
\end{center}
\vskip-3mm
\caption{Same as in Fig. \ref{fig::subsets-F-gens}, except that CDFs for \textbf{impostor comparisons} are presented, together with respective mean values provided in brackets.}
\label{fig::subsets-F-imps}
\end{figure}

\begin{table}[!htb]
\renewcommand{\arraystretch}{1.1}
\caption{Kolmogorov-Smirnov statistical testing results for the disease influence type experiment, genuine comparisons. The null hypotheses $H_0$ in all tests state that the samples originating from two compared partitions are drawn from the same distribution. Alternative hypotheses are detailed in rows labeled $H_1$. One-sided test is used. $F(g_k)$ denotes the cumulative distribution function of $g_k$, where $g_k$ denotes the genuine scores calculated to the $k$-th partition.}
\label{table:statTestsGenuine}
\centering\footnotesize
\begin{tabular}[t]{cc|c|c|c|c|}
\cline{3-6}
& & \emph{Clear} ($g_c$) & \emph{Geometry} ($g_g$) & \emph{Tissue} ($g_t$) & \emph{Obstructions} ($g_o$) \\
& & vs. \emph{Healthy} ($g_h$) & vs. \emph{Healthy} ($g_h$) & vs. \emph{Healthy} ($g_h$) & vs. \emph{Healthy} ($g_h$) \\\hline\hline
\multicolumn{1}{|c}{\multirow{2}{*}{\bf OSIRIS}} & \multicolumn{1}{|c|}{$H_1$} & $F(g_c) < F(g_h)$ & $F(g_g) < F(g_h)$ & $F(g_t) < F(g_h)$ & $F(g_o) < F(g_h)$  \\\cline{2-6}
\multicolumn{1}{|c}{} & \multicolumn{1}{|c|}{$p$-value} & $<0.0001$  & $<0.0001$ & \textcolor{red}{0.1071}  & $<0.0001$ \\\hline
\multicolumn{1}{|c}{\multirow{2}{*}{\bf MIRLIN}} & \multicolumn{1}{|c|}{$H_1$} & $F(g_c) < F(g_h)$ & $F(g_g) < F(g_h)$ & $F(g_t)<F(g_h)$ & $F(g_o)<F(g_h)$  \\\cline{2-6}
\multicolumn{1}{|c}{} & \multicolumn{1}{|c|}{$p$-value} & $<0.0001$  & $<0.0001$ & 0.0039 & $<0.0001$ \\\hline
\multicolumn{1}{|c}{\multirow{2}{*}{\bf IriCore}} & \multicolumn{1}{|c|}{$H_1$} & $F(g_c) < F(g_h)$ & $F(g_g) < F(g_h)$ & $F(g_t)<(g_h)$ & $F(g_o)<F(g_h)$  \\\cline{2-6}
\multicolumn{1}{|c}{} & \multicolumn{1}{|c|}{$p$-value} & $<0.0001$ & $<0.0001$ & \textcolor{red}{0.3325} & $<0.0001$ \\\hline
\multicolumn{1}{|c}{\multirow{2}{*}{\bf VeriEye}} & \multicolumn{1}{|c|}{$H_1$} & $F(g_c) > F(g_h)$ & $F(g_g) > F(g_h)$ & $F(g_t) > F(g_h)$ & $F(g_o) > F(g_h)$  \\\cline{2-6}
\multicolumn{1}{|c}{} & \multicolumn{1}{|c|}{$p$-value} & $<0.0001$ & $<0.0001$ & \textcolor{red}{0.6942} & $<0.0001$ \\\hline
\end{tabular}
\end{table}

\begin{table}[!htb]
\renewcommand{\arraystretch}{1.1}
\caption{Same as in Table \ref{table:statTestsGenuine}, except that impostor comparison scores are analyzed and two-sided Kolmogorov-Smirnov test (for the resampled data) was applied. $F(i_k)$ denotes the cumulative distribution function of $i_k$, where $i_k$ denotes the impostor scores calculated to $k$-th partition.}
\label{table:statTestsImpostor}
\centering\footnotesize
\begin{tabular}[t]{c|c|c|c|c|}
\cline{2-5}
& \emph{Clear} ($i_c$) & \emph{Geometry} ($i_g$) & \emph{Tissue} ($i_t$) & \emph{Obstructions} ($i_o$) \\
& vs. \emph{Healthy} ($i_h$) & vs. \emph{Healthy} ($i_h$) & vs. \emph{Healthy} ($i_h$) & vs. \emph{Healthy} ($i_h$) \\
& $H_1: F(i_c) \nsim F(i_h)$ & $H_1: F(i_g) \nsim F(i_h)$ & $H_1: F(i_t) \nsim F(i_h)$ &  $H_1: F(i_o) \nsim F(i_h)$\\\hline\hline
\multicolumn{1}{|c|}{\bf OSIRIS} & $<0.0001$ & $<0.0001$ & $<0.0001$ & $<0.0001$ \\\hline
\multicolumn{1}{|c|}{\bf MIRLIN} & 0.0078 & $<0.0001$ & $<0.0001$ & $<0.0001$ \\\hline
\multicolumn{1}{|c|}{\bf IriCore} & $<0.0001$ & $<0.0001$ & $<0.0001$ & $<0.0001$ \\\hline
\multicolumn{1}{|c|}{\bf VeriEye} & 0.0051 & $<0.0001$ & $<0.0001$ & $<0.0001$ \\\hline
\end{tabular}
\end{table}

\begin{figure}[htb!]
\begin{center}
	\includegraphics[width=0.497\linewidth]{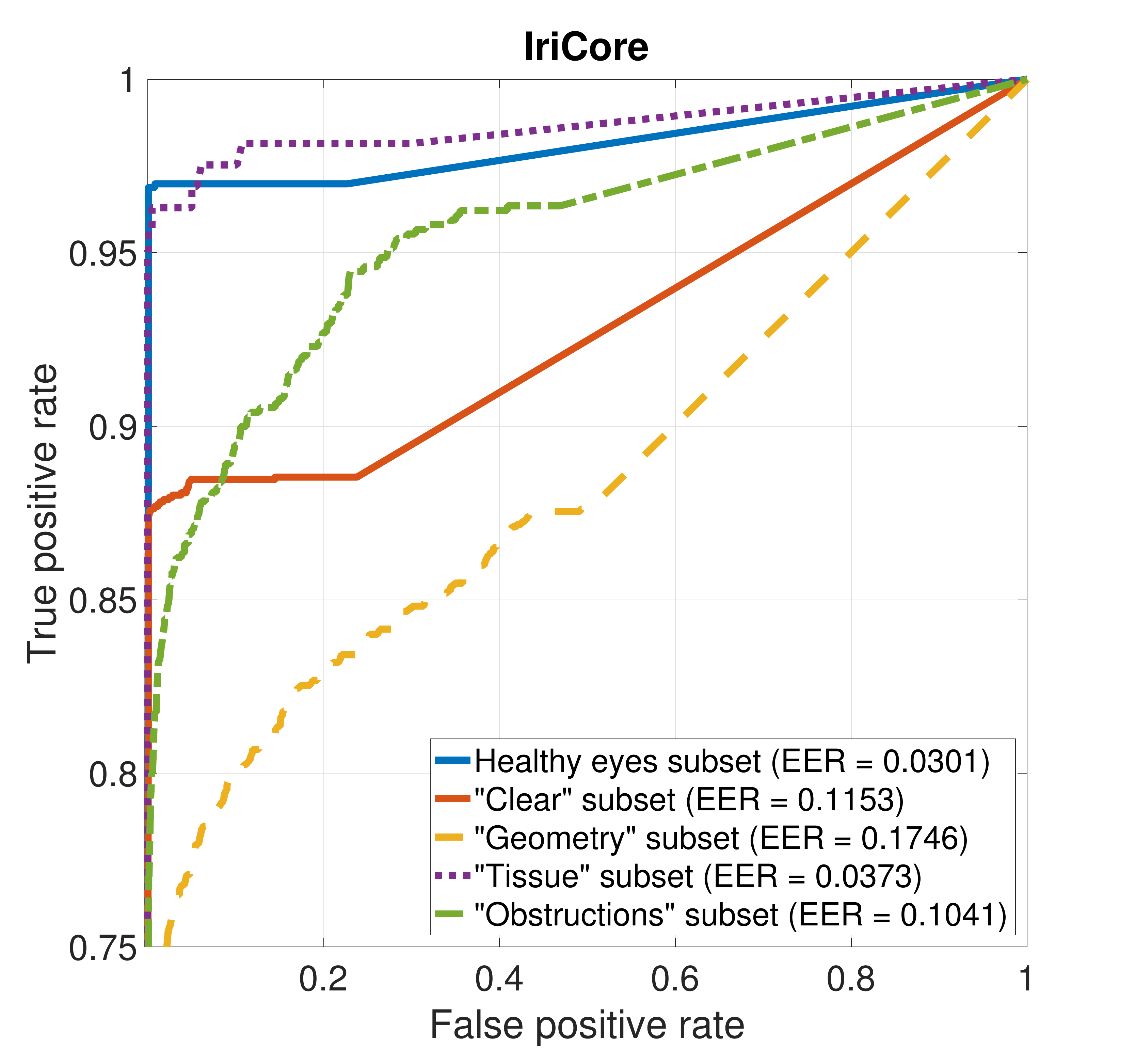}
	\hfill
	\includegraphics[width=0.497\linewidth]{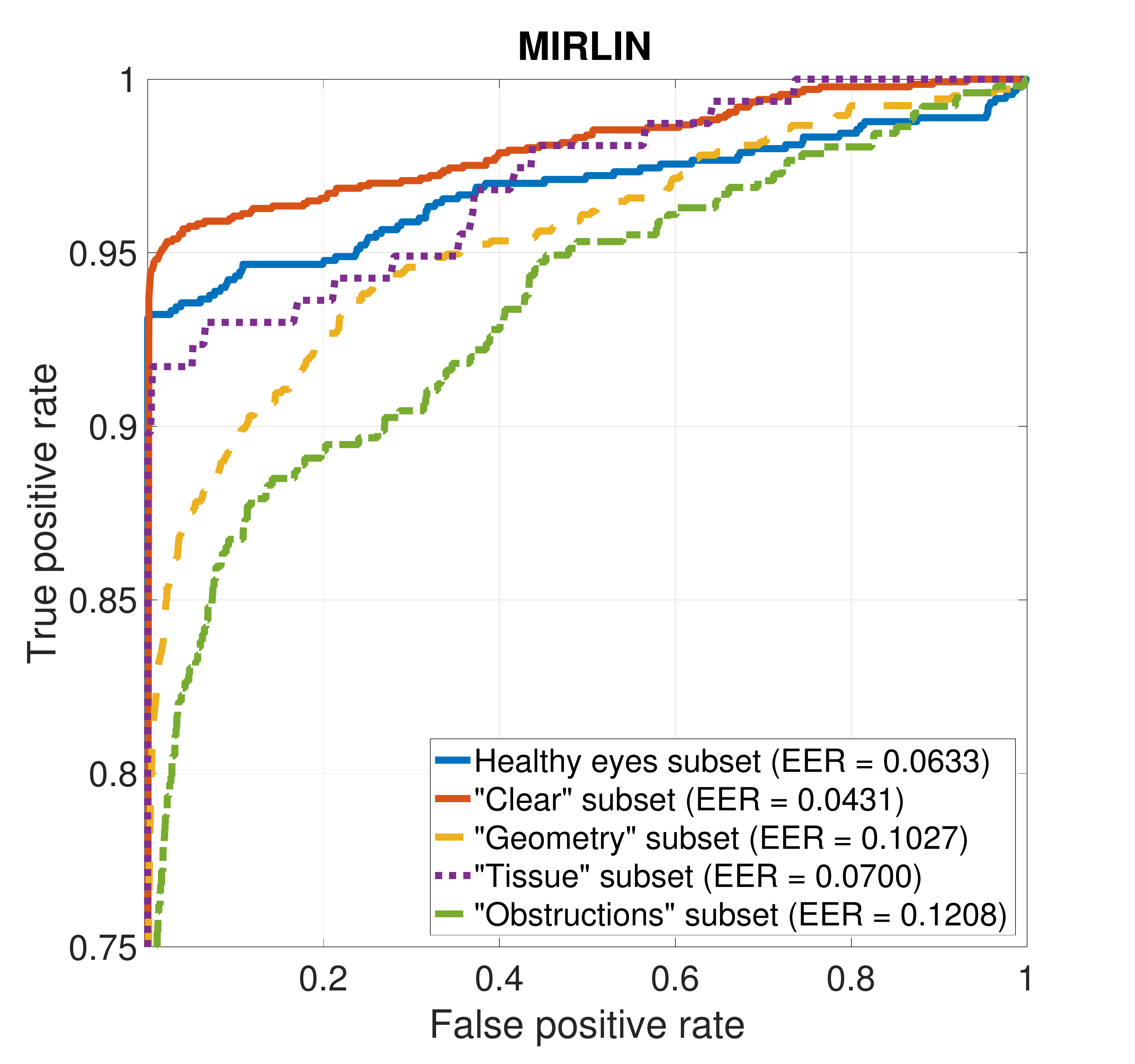}
	\vfill
	 \includegraphics[width=0.497\linewidth]{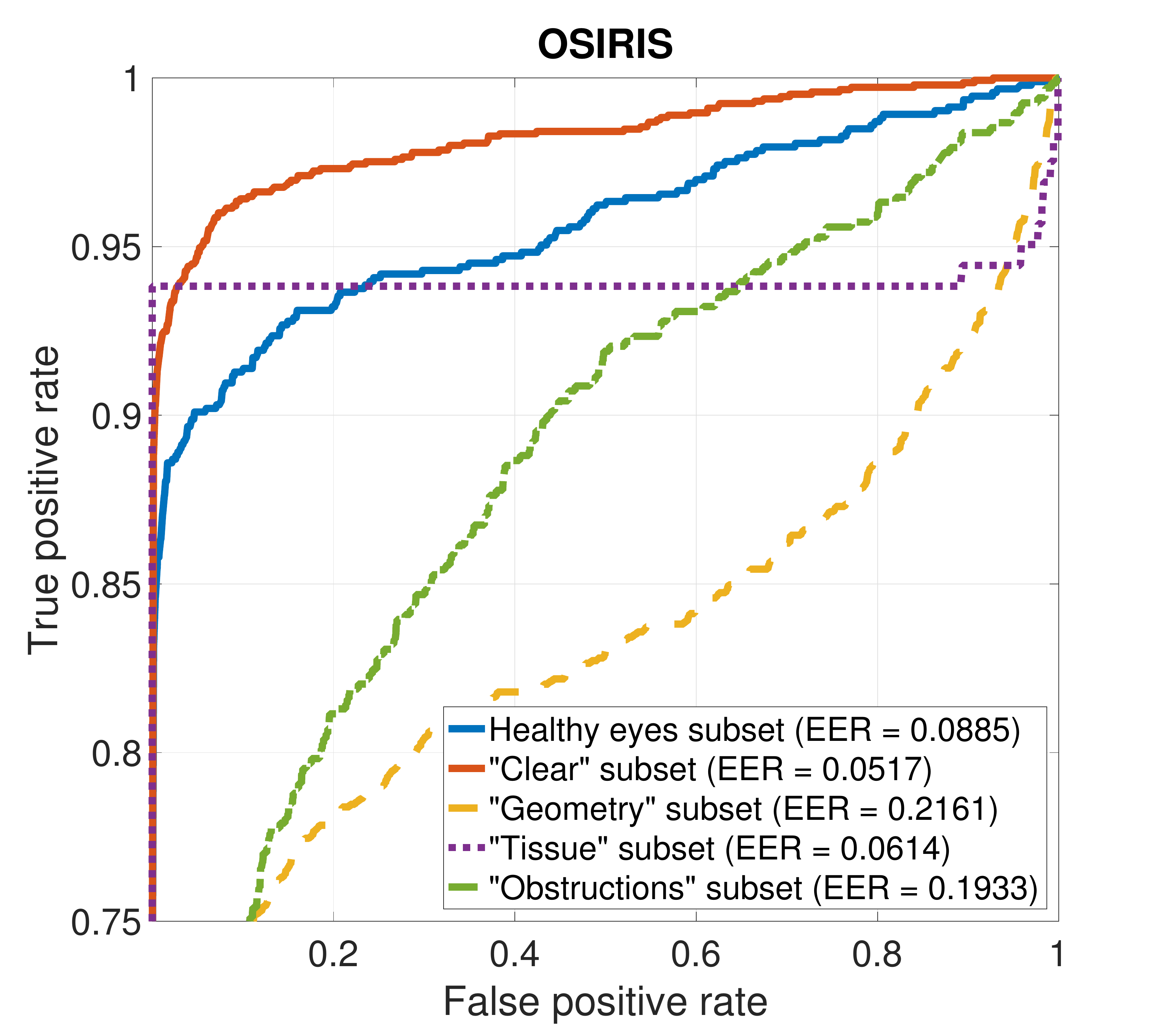}
  \hfill
  \includegraphics[width=0.497\linewidth]{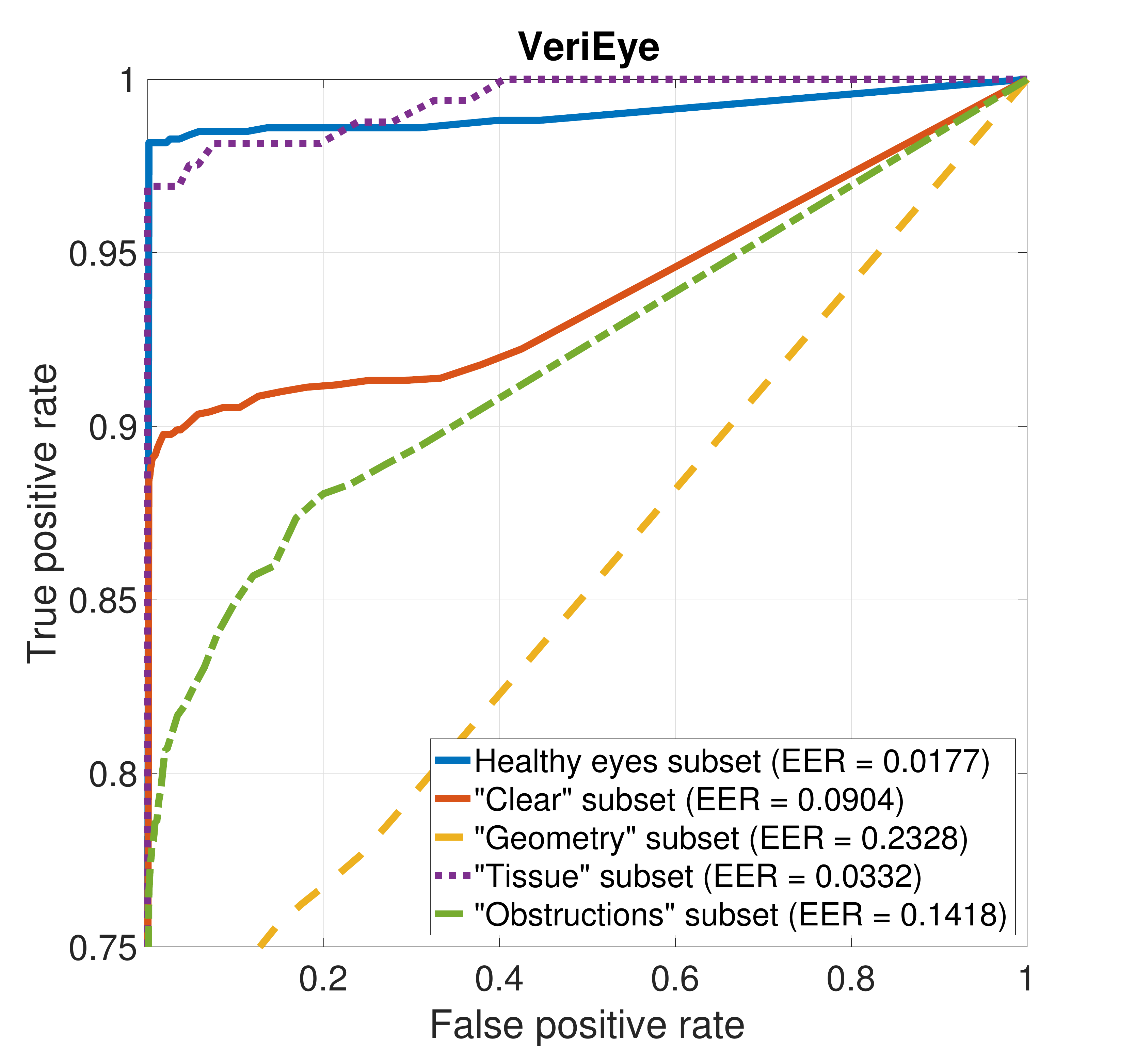}
\end{center}
\vskip-3mm
\caption{Receiver Operating Characteristic (ROC) curves obtained for all four iris recognition methods denoting the performance of these systems for five data subsets. Equal Error Rate values are provided in brackets.}
\label{fig::subsets-ROCs}
\end{figure}

Fig. \ref{fig::subsets-ROCs} present ROC curves plotted collectively for all five data subsets representing different types of damage inflicted to the eye. This is repeated for all four of the iris matchers involved in this study. Here as well, the \emph{Geometry} and \emph{Obstructions} subsets are giving the worst performance. Surprisingly, for the IriCore and VeriEye matchers, the \emph{Clear} subset also performs poorly, while the \emph{Tissue} subset is behaving similarly or better than the \emph{Healthy} subset.

\subsection{Recognition errors analysis} 
As finding the actual reasons behind erroneous performance is crucial for getting a complete picture of the studied phenomenon, we performed a careful visual inspection of the samples that generated exceptionally poor comparison scores. As impostor comparison scores are not impacted in a significant way, this is done only for the genuine scores. Since bad performance in iris recognition typically originates from incorrect execution of the segmentation stage, we employed two of the iris matchers that are capable of showing image segmentation results: MIRLIN and OSIRIS, to generate iris images with denoted iris localization results. This analysis confirmed that failed iris localization is the most prevalent source of bad iris matcher performance. Segmentation errors that we have come across were most likely caused by some artifacts, such as distortions in the pupil boundary, obstructions such as corneal hazes, or damages to the iris tissue being interpreted by image segmentation algorithms as the pupil itself. Thus, the following matching stage, which is executed after the segmentation stage, could not be performed correctly, but instead was performed using the non-iris portions of the image. This is especially true for \emph{Geometry} and \emph{Obstructions} subsets of the data, which is coherent with exceptionally poor ROC-wise performance of the data belonging to these subsets. VeriEye and IriCore algorithms do not provide a way to read the segmentation results, however, an examination of those samples that perform the worst when using these method, easily identifies conditions responsible for errors, namely: significant geometrical distortions, severe corneal hazes, blurred boundary between the iris and the pupil, letting us hazard a guess that segmentation issues are the ones responsible for errors here as well.


\section{Conclusions}
\label{sec:Conslusions}
This book chapter summarizes the Authors' knowledge regarding iris recognition behavior under conditions involving ophthalmic disorders, including both mild illnesses and severe eye pathologies. Together with an extensive literature review concerning this subject, two extensive experiments are described with important results delivered.

\begin{svgraybox}
	The first experiment related to probably the most proliferated eye illness worldwide, the cataract, proves that despite usually not affecting the eye and the iris in a significant way, this pathology is capable of causing serious negative impact on the performance of state-of-the-art iris recognition technologies used today. With Equal Error Rates for the cataract-affected eyes being a few percent higher than those obtained with data corresponding to healthy eyes, one may expect recognition accuracy to be noticeably lower for people suffering from this illness. Combining this with high number of cataract occurrences, especially in third-world countries, leads to a conclusion that this issue should be seriously taken into consideration when building future, large-scale biometric applications employing iris recognition. 
\end{svgraybox}

\begin{svgraybox}
	The latter of the two experiments deals with a more broad and universal approach to the problem and tries to predict the recognition accuracy deterioration with respect to the type of damage inflicted by pathological processes in the eye, regardless of the actual medical origin and disease taxonomy. Ophthalmological disorders that are expected to cause the highest performance drops are those causing the pupil to be irregularly shaped and those introducing obstructions that make correct imaging of the iris texture difficult or impossible. An attempt to explain the underlying reasons of such poor performance is carried out, pointing to image segmentation errors as the predominant source of performance deterioration. However, knowing which types of eye damage are the ones most likely to cause recognition errors, one can employ visual inspection of the eyes of a person under enrollment to assess whether iris recognition can be reliably used to manage this person's identity.
\end{svgraybox}

With such knowledge in place, more research in this field with more data is necessary to better quantify the effect ophthalmic disorders have on iris recognition systems on larger scales and to be able to propose appropriate countermeasures against the reported drops in recognition accuracy to make this biometric technology an even better and more reliable solution to use globally without excluding subjects suffering from eye illnesses.

\begin{acknowledgement}
All patients who participated in this study have been provided with detailed information regarding the research and an informed, written consent has been obtained from each volunteer. Ethical principles of the Helsinki Declaration were carefully followed by the authors.   
\end{acknowledgement}

\bibliographystyle{ieee} 
 
\small
\bibliography{refs}

\end{document}